\documentclass[11pt,a4paper]{article}
\usepackage{authblk}
\usepackage[hyperref]{acl2020}
\usepackage{times}
\usepackage{latexsym}

\usepackage{graphicx}
\usepackage{subfigure}
\usepackage{url}

\usepackage{amsmath,amsfonts,bm}

\def\eqref#1{equation~\ref{#1}}

\def\1{\bm{1}}

\def\vi{{\bm{i}}}

\DeclareMathAlphabet{\mathsfit}{\encodingdefault}{\sfdefault}{m}{sl}
\SetMathAlphabet{\mathsfit}{bold}{\encodingdefault}{\sfdefault}{bx}{n}

\DeclareMathOperator*{\argmax}{arg\,max}

\DeclareMathSymbol{*}{\mathbin}{symbols}{"03}
\DeclareMathSymbol{\ast}{\mathbin}{symbols}{"03}

\usepackage{array}
\usepackage{ltablex,booktabs}
\usepackage{makecell}
\usepackage{todonotes}
\usepackage{xr} 
\usepackage{multirow}
\usepackage{tikz}
\usepackage{diagbox}

\usepackage{xpatch}
\xpatchcmd{\author}{\relax#1\relax}{\relax\detokenize{#1}\relax}{}{}

\usepackage{microtype}

\aclfinalcopy 
\def\aclpaperid{512} 

\title{Compositionality and Generalization in Emergent Languages}

\author[\empty]{Rahma Chaabouni\textsuperscript{1,2}\thanks{\hspace{0.1cm}Contributed equally.}}
\newcommand\CoAuthorMark{\footnotemark[\arabic{footnote}]} 
\author[\empty]{Eugene Kharitonov\textsuperscript{1}\protect\CoAuthorMark}
\author[1]{Diane Bouchacourt}
\author[1,2]{Emmanuel Dupoux}
\author[1,3]{Marco Baroni}
\affil[1]{Facebook AI Research}
\affil[2]{Cognitive Machine Learning (ENS - EHESS - PSL Research University - CNRS - INRIA)}
\affil[3]{ICREA}
\affil[ ]{\tt {\{rchaabouni,kharitonov,dianeb,dpx,mbaroni\}@fb.com}}

\date{}

\begin{document}
\maketitle
\begin{abstract}
Natural language allows us to refer to novel composite concepts by combining expressions denoting their parts according to systematic rules, a property known as \emph{compositionality}. In this paper, we study whether the language emerging in deep multi-agent simulations possesses a similar ability to refer to novel primitive combinations, and whether it accomplishes this feat by strategies akin to human-language compositionality. Equipped with new ways to measure compositionality in emergent
languages inspired by disentanglement in representation learning, we
establish three main results. First, given sufficiently large input
spaces, the emergent language will naturally develop the ability to
refer to novel composite concepts. Second, there is no correlation
between the degree of compositionality of an emergent language and its
ability to generalize. Third, while compositionality is not necessary
for generalization, it provides an advantage in terms of language
transmission: The more compositional a language is, the more easily it
will be picked up by new learners, even when the latter differ in
architecture from the original agents. We conclude that
compositionality does not arise from simple generalization pressure,
but if an emergent language does chance upon it, it will be more
likely to survive and thrive.
\end{abstract}

\section{Introduction}

Most concepts we need to express are composite in some way. Language
gives us the prodigious ability to assemble messages
referring to novel composite concepts by systematically combining
expressions denoting their parts. As
interest raises in developing deep neural agents evolving a
communication code to better accomplish cooperative tasks, the
question arises of how the emergent code can be endowed with the same
desirable \emph{compositionality} property
\cite{Kottur:etal:2017,Lazaridou:etal:2018,mordatch2018emergence,cogswell2019,Li2019}. This in turn requires measures of how compositional an emergent
language is \cite{Andreas:2019}. Compositionality is a core notion in linguistics \cite{Partee:2004}, but linguists' definitions assume full knowledge of primitive expressions
and their combination rules, which we lack when analyzing emergent
languages \cite{Nefdt:2020}. Also, these definitions are categorical, whereas to compare
emergent languages we need to quantify  degrees of
compositionality.

Some researchers equate compositionality with the ability to correctly
refer to unseen composite inputs
\cite[e.g.,][]{Kottur:etal:2017,cogswell2019}. This approach  measures the generalization ability of a
language, but it does not provide any insights on \emph{how} this ability comes about. Indeed, one of our main results below
is that emergent languages can attain perfect generalization without
abiding to intuitive notions of compositionality.

Topographic similarity has become the standard way to quantify the
compositionality of emergent languages
\cite[e.g.,][]{Brighton2006,Lazaridou:etal:2018, Li2019}. This metric measures whether the distance between two meanings
correlates with the distance between the messages expressing
them. While more informative than generalization, topographic
similarity is still rather agnostic about the nature of
composition. For example, when using, as is standard practice,
Levenshtein distance to measure message distance, an emergent language
transparently concatenating symbols in a fixed order and one mixing
deletion and insertion operations on free-ordered symbols can have the
same topographic similarity. 

We introduce here two more ``opinionated'' measures of
compositionality that capture some intuitive properties of what we would expect to happen in a compositional emergent language. %
One possibility we consider is that order-independent juxtapositions
of primitive forms could denote the corresponding union of meanings,
as in English noun conjunctions: \emph{cats and dogs}, \emph{dogs and
  cats}. The second still relies on juxtaposition, but exploits order
to denote different classes of meanings, as in English adjective-noun
phrases: \emph{red triangle}, \emph{blue square}. Both strategies
result in \emph{disentangled} messages, where each primitive symbol
(or symbol+position pair) univocally refers to a distinct primitive meaning
independently of context. We consequently take inspiration from work
on disentanglement in representation learning \cite{Suter2019} to
craft measures that quantify whether an emergent language follows one
of the proposed composition strategies.

Equipped with these metrics, we proceed to ask the following
questions. First, are neural agents able to generalize
to unseen input combinations in a simple communication game? We find
that generalizing languages reliably emerge when the input domain is
sufficiently large. This somewhat expected result is important
nevertheless, as failure-to-generalize claims in the recent literature
are often based on very small input spaces. Second, we unveil a
complex interplay between compositionality and generalization. On the
one hand, there is no correlation between our compositionality metrics
and the ability to generalize, as emergent languages
successfully refer to novel composite concepts in inscrutablly
entangled ways. (Order-dependent) compositionality, however, if not
necessary, turns out to be a sufficient condition for
generalization. Finally, more compositional languages are easier
to learn for new agents, including agents that are architecturally
different from the ones that evolved the language. %
This suggests that, while composition might not be a
``natural'' outcome of the need to generalize, it is a highly
desirable one, as compositional languages will more easily be adopted
by a large community of different agents. We return to the
implications of our findings in the discussion.

\section{Setup}
\subsection{The game}

We designed a variant of Lewis' signaling game \cite{Lewis:1969}. The game proceeds as follows:

\begin{enumerate}
    \item Sender network receives one input $\vi$ and chooses a sequence of symbols from its vocabulary $V=\{s_1, s_2..., s_{c_{voc}}\}$ of size $c_{voc}$ to construct a message $m$ of fixed length $c_{len}$.
    \item Receiver network consumes $m$ and outputs $\hat{\vi}$.
    \item Agents are successful if $\hat{\vi}=\vi$, that is, Receiver reconstructs Sender's input.
\end{enumerate}

Each input $\vi$ of the reconstruction game is comprised of $i_{att}$
attributes, each with $i_{val}$ possible values. We let $i_{att}$
range from $2$ to $4$ and $i_{val}$ from $4$ to $100$.  We represent each attribute as a $i_{val}$
one-hot vector. An input $\vi$ is given by the concatenation of its attributes.
For a given
($i_{att}$, $i_{val}$), the number of input samples
$ |I|=i_{val}^{i_{att}}$.

This environment, which can be seen as an extension of that of
\citet{Kottur:etal:2017}, is one of the simplest possible settings to
study the emergence of reference to composite concepts (here,
combinations of multiple attributes). Attributes can be seen as
describing object properties such as color and shape, with their
values specifying those properties for particular objects (\emph{red},
\emph{round}). Alternatively, they could be seen as slots in an
abstract semantic tree (e.g., agent and action), with the values
specifying their fillers (e.g., \emph{dog}, \emph{barking}). In the
name of maximally simplifying the setup and easing interpretability,
unlike \citet{Kottur:etal:2017}, we consider a single-step game. We
moreover focus on input reconstruction instead of discrimination of a
target input among distractors as the latter option adds furtherx
complications: for example, languages in that setup have been shown to be
sensitive to the number and distribution of the distractors
\cite{Lazaridou:etal:2018}.

For a fixed $|I|$, we endow Sender with large enough channel capacity $|C|=c_{voc}^{c_{len}}$ ($c_{voc} \in \{5,10,50,100\}$ and $c_{len} \in \{3,4,6,8\}$) to express the whole input space (i.e., $|C| \geq |I|$). Unless explicitly mentioned, we run $10$ different initializations per setting. See Appendix \ref{sec:gridSearch} for details about the range of tested settings. %
%
The game is implemented in EGG \cite{Kharitonov:etal:2019}.\footnote{Code can be found at \url{https://github.com/facebookresearch/EGG/tree/master/egg/zoo/compo_vs_generalization}.}

\subsection{Agent architecture}
\label{ssArch}
Both agents are implemented as single-layer GRU cells~\citep{Cho2014} with hidden states of size 500.\footnote{Experiments with GRUs of different capacity are reported in the Appendix. We also informally replicated our main results with
LSTMs, that were slower to converge. We were unable to adapt Transformers to successfully play our game.} %
Sender encodes $\vi$ in a message $m$ of fixed length $c_{len}$ as follows. First, a linear layer maps the input vector into the initial hidden state of Sender. Next, the message is generated symbol-by-symbol by sampling from a Categorical distribution over the vocabulary $c_{voc}$, parameterized by a linear mapping from Sender's hidden state.  The generated symbols are fed back to the cell.  At test time, instead of sampling, symbols are selected greedily.


Receiver consumes the entire message $m$. Further, we pass its hidden state through a linear layer and consider the resulting vector as a concatenation of $i_{att}$ probability vectors over $i_{val}$ values each. As a loss, we use the average cross-entropy between these distributions and Sender's input.

\subsection{Optimization}
Popular approaches for training with discrete communication include Gumbel-Softmax~\cite{Maddison2016,Jang2016}, REINFORCE~\cite{Williams1992}, and a hybrid in which the Receiver gradients are calculated via back-propagation and those of Sender via REINFORCE \citep{Schulman2015}. We use the latter, as recent work \cite[e.g.,][]{Chaabouni:etal:2019} found it to converge more robustly. 
We apply standard tricks to improve convergence: (a) running mean baseline to reduce the variance of the gradient estimates~\cite{Williams1992}, and (b) a term in the loss that favors higher entropy of Sender's output, thus promoting exploration. The obtained gradients are passed to the Adam optimizer~\citep{Kingma2014} with learning rate $0.001$.

\section{Measurements}
\subsection{Compositionality}
\label{sMeasuring}
\textbf{Topographic similarity (topsim)}~\cite{Brighton2006} is commonly used in language emergence studies as a quantitative proxy for compositionality~\cite[e.g.,][]{Lazaridou:etal:2018, Li2019}. Given a distance function in the input space (in our case, attribute value overlap, as attributes are unordered, and values categorical) and a distance function in message space (in our case, following standard practice, minimum edit distance between messages), \emph{topsim} is the (Spearman) correlation between pairwise input distances and the corresponding message distances. The measure can detect a tendency for messages with similar meanings to be similar in form, but it is relatively agnostic about the type of similarity (as long as it is captured by minimum edit distance).

We complement \emph{topsim} with two measures that probe for more
specific types of compositionality, that we believe capture what
deep-agent emergent-language researchers seek for, when interested in
compositional languages. In most scenarios currently considered in
this line of research, the composite inputs agents must refer to are
sets or sequences of primitive elements: for example, the values of a
set of attributes, as in our experiment. In this restricted setup, a compositional language
is a language where symbols independently referring to primitive input
elements can be juxtaposed to jointly refer to the input
ensembles. Consider a language with a symbol \emph{r} referring to
input element \emph{color:red} and another symbol \emph{l} referring
to \emph{weight:light}, where \emph{r} and \emph{l} can be juxtaposed
(possibly, in accordance with the syntactic rules of the language) to
refer to the input set $\{$\emph{color:red}, \emph{weight:light}$\}$. This language
is intuitively compositional. On the other hand, a language where both
\emph{r} and \emph{l} refer to these two input elements, but only when
used together, whereas other symbol combinations would refer to
\emph{color:red} and \emph{weight:light} in other contexts, is
intuitively not compositional. Natural languages support forms
of compositionality beyond the simple juxtaposition of
context-independent symbols to denote ensembles of input elements we are considering here
(e.g., constructions that denote the application of functions to
arguments). However, we believe that the proposed intuition is
adequate for the current state of affairs in language emergence research.

The view of compositionality we just sketched is closely
related to the idea of disentanglement in representation learning. Disentangled representations are expected to enable a consequent model to generalize on new domains and tasks \citep{Bengio2013}. Even if this claim has been challenged \citep{Bozkurt2019, Locatello2019},  several interesting metrics have been proposed to quantify disentanglement, as reviewed in \citet{Suter2019}. %
We build in particular upon the \textit{Information Gap} disentanglement measure of \citet{Chen2018}, evaluating how well representations capture independence in the input sets.

Our \textbf{positional disentanglement (posdis)} metric 
 measures whether symbols \emph{in specific positions} tend to univocally refer to the values of a specific attribute. This order-dependent strategy is commonly encountered in natural language structures (and it is a pre-condition for sophisticated syntactic structures to emerge). Consider English adjective-noun phrases with a fully intersective interpretation, such as \emph{yellow triangle}. Here, the words in the first slot will refer to adjectival meanings, those in the second to nominal meanings. In our simple environment, it might be the case that the first symbol is used to discriminate among values of an attribute, and the second to discriminate among values of another attribute.
Let's denote $s_j$ the $j$-th symbol of a message and
 $a_1^j$ the attribute that has the highest mutual  information with $s_j$: $a^j_1 = \argmax_a \mathcal{I}(s_j; a)$. In turn, $a_2^j$ is the second highest informative attribute, $a^j_2 = \argmax_{a \neq a_1^j} \mathcal{I}(s_j; a)$. Denoting $\mathcal{H}(s_j)$ the entropy of $j$-th position (used as a normalizing term), we define \textit{posdis} as:
\begin{equation}
\label{eq:posdis}
posdis = 1/c_{len} \sum_{j=1}^{c_{len}} \frac{\mathcal{I}(s_j; a^j_1) - \mathcal{I}(s_j; a^j_2)}{\mathcal{H}(s_j)}
\end{equation}
We ignore positions with zero entropy. Eq.~\ref{eq:posdis} captures the intuition that, for a language to be compositional given our inputs, each position of the message should only be informative about a single attribute. However, unlike the related measure proposed by \citet{Resnick2019}, it does not require knowing which set of positions encodes a particular attribute, which makes it computationally simpler (only linear in $c_{len}$).

\emph{Posdis} assumes that a language uses positional information to disambiguate symbols. However, we can easily imagine a language where symbols univocally refer to distinct input elements independently of where they occur, making order irrelevant.\footnote{This is not unlike what happens in order-insensitive constructions such as English conjunctions: \emph{dogs and cats}, \emph{cats and dogs}.} Hence, we also introduce \textbf{bag-of-symbols disentanglement (bosdis)}. The latter maintains the requirement for symbols to univocally refer to distinct meanings, but captures the intuition of a permutation-invariant language, where only symbol counts are informative. Denoting by $n_j$ a counter of the $j$-th symbol in a message, \textit{bosdis} is given by:
\begin{equation}
bosdis = 1/c_{voc} \sum_{j=1}^{c_{voc}} \frac{\mathcal{I}(n_j; a^j_1) - \mathcal{I}(n_j; a^j_2)}{\mathcal{H}(n_j)}
\end{equation}
In all experiments, the proposed measures \emph{topsim}, \emph{posdis} and \emph{bosdis} are calculated on the train set.

In Appendix \ref{sec:minLang}, we illustrate how the three metrics behave differently on three miniature languages. Across the languages of all converging runs in our simulations, their Spearman correlations are: \emph{topsim}/\emph{posdis}: $0.08$; \emph{topsim}/\emph{bosdis}: $0.38$; \emph{posdis}/\emph{bosdis}: $0.31$. These correlations, while not extremely high, are statistically significant ($p<0.01$), which is reassuring as all metrics attempt to capture compositionality.
It is also in line with reasonable expectations that the most ``opinionated'' \emph{posdis} measure is the one that behaves most differently from \emph{topsim}.



\subsection{Generalization}
In our setup, generalization can be straightforwardly measured by splitting all possible distinct inputs so that the test set only contains inputs with attribute combinations that were not observed at training. Generalization is then simply quantified by test accuracy. In intuitive terms, at training time the agents are exposed to \emph{blue triangles} and \emph{red circles}, but \emph{blue circles} only appear at test time. This requires Sender to generate new messages, \emph{and} Receiver to correctly infer their meaning. If a \emph{blue circle} is  accurately reconstructed, then agents do generalize.

For all the considered settings, we split the possible distinct inputs into $90\%$ train and $10\%$ test items. This implies that the absolute training/test set sizes increase with input dimension (this issue is further discussed in Appendix \ref{sec:density}).

Finally, we only evaluate generalization for runs that successfully converged, where convergence is operationalized as $>99.9\%$ training-set accuracy.


\section{Generalization emerges ``naturally" if the input space is large}
\label{sGeneralization}

\begin{figure}
   \centering
      \includegraphics[width=0.8\columnwidth]{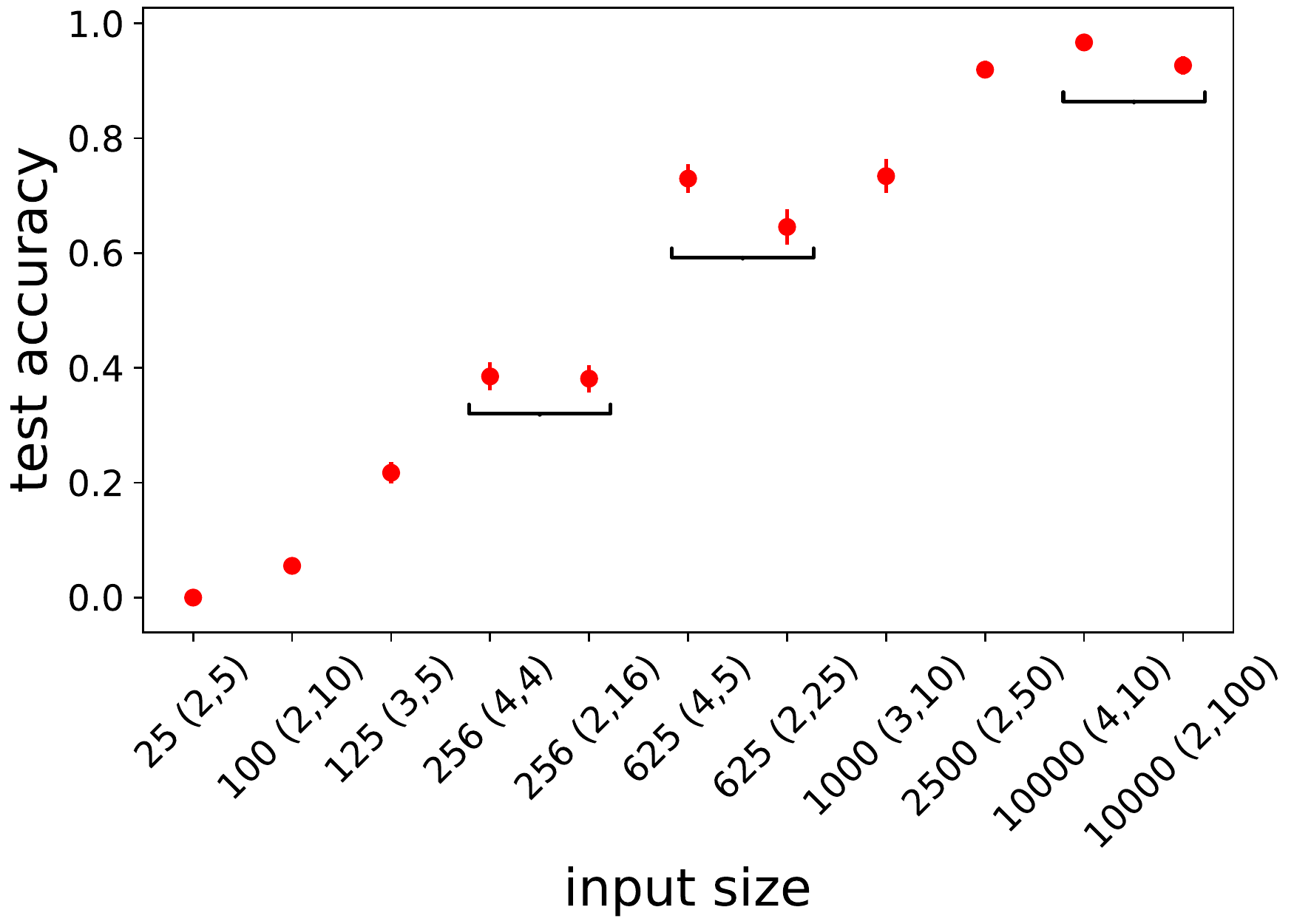}
      \caption{Average accuracy on unseen combinations as a function
        of input size of successful runs. The x-axis is ordered
        by increasing input size $|I|$. Brackets denote
        ($i_{att}$,
        $i_{val}$). 
        Vertical bars represent the standard error of the mean (SEM). Horizontal
        brackets group settings with same $|I|$ but different
        ($i_{att}$, $i_{val}$).}
\label{fig:generalization1}
\end{figure}


Fig.~\ref{fig:generalization1} shows that emergent languages are
able to almost perfectly generalize to unseen combinations \emph{as
  long as input size} $|I|$ is sufficiently large (input size/test
accuracy Spearman $\rho=0.86$, $p\approx{}0$). The figure also shows
that the \emph{way} in which a large input space is obtained
(manipulating $i_{att}$ or $i_{val}$) does not matter (no significant
accuracy difference between the bracketed runs, according to a set of
t-tests with $p>0.01$). Moreover, the correlation is robust to varying
agents' capacity (Appendix \ref{sec:robust}; see \citet{Resnick2019}
for a thorough study of how agent capacity impacts generalization and
compositionality). Importantly, the effect is not simply a product of
larger input sizes coming with larger training corpora, as we replicate
it in Appendix \ref{sec:density} while keeping the number of
distinct training examples fixed, but varying input
\emph{combinatorial variety}. What matters is that, in the training
data, specific attribute values tend to occur with a large range
of values from other attributes, providing a cue about the composite
nature of the input.

That languages capable to generalize will only emerge when the input
is varied enough might seem obvious, and it has been shown before in
mathematical simulations of language emergence
\cite{nowak2000evolution}, as well as in studies of deep network
inductive biases \cite{zhao2018}. However, our result suggests an
important \emph{caveat} when interpreting experiments  based on small input environments that report
failures in the generalization abilities of deep networks
\cite[e.g.,][]{Kottur:etal:2017,Lake:Baroni:2017}. Before assuming
that special architectures or training methods are needed for
generalization to emerge, such experiments should be repeated with
much larger/varied input spaces, where it is harder for agents to
develop ad-hoc strategies overfitting the training data and failing to generalize.

We also considered the relation between channel capacity $|C|$ and
language emergence. Note that $|C|\geq|I|$ is a prerequisite for
successful communication, and a perfectly compositional language could
already generalize at the lower $|C|=|I|$ bound. Indeed, limiting
channel capacity has been proposed as an important constraint for the
emergence of compositionality \cite{Nowak:Krakauer:1999}. However, we
find that, when $|I|$ is sufficiently large to support generalization,
our deep agents need $|C|>|I|$ in order to even converge at training
time. The \emph{minimum} $|C|/|I|$ ratio across all converging runs
for each configuration with $|I|\geq 625$ (the settings where we witness generalizing languages) is on average 5.9 (s.d.:
4.4). 

Concretely, this implies that none of our successful languages is as compact as a minimal fully-compositional
solution would afford. Appendix \ref{sec:channelgen} reports experiments focusing, more
specifically, on the relation between channel capacity and
generalization, showing that it is essential for $|C|$ to be above a
large threshold to reach near-perfect accuracy, and further increasing
$|C|$ beyond that does not hamper generalization.


\section{Generalization does not require compositionality}
\label{sGeneralizationCompositionality}

Having established that emergent languages \emph{can} generalize to
new composite concepts, we test whether languages that generalize
better are also more compositional. Since \emph{bosdis} and
\emph{topsim} correlate with $|C|$ (Appendix \ref{sec:dependence}), we compute
Spearman correlations between test accuracy and compositionality
metrics across all converging runs of each ($i_{att}$, $i_{val}$,
$c_{len}$, $c_{voc}$) configuration separately. Surprisingly, in just 4 out
of $141$ distinct settings the correlation is significant ($p<0.01$)
for at least 1 measure.\footnote{$3$, $3$ and $1$ (different)
  significant settings for \textit{topsim}, \textit{posdis} and
  \textit{bosdis}, respectively.}

\begin{figure*}[ht]
\centering
\subfigure[\textit{posdis}]{
    \includegraphics[width=0.3\textwidth, keepaspectratio]{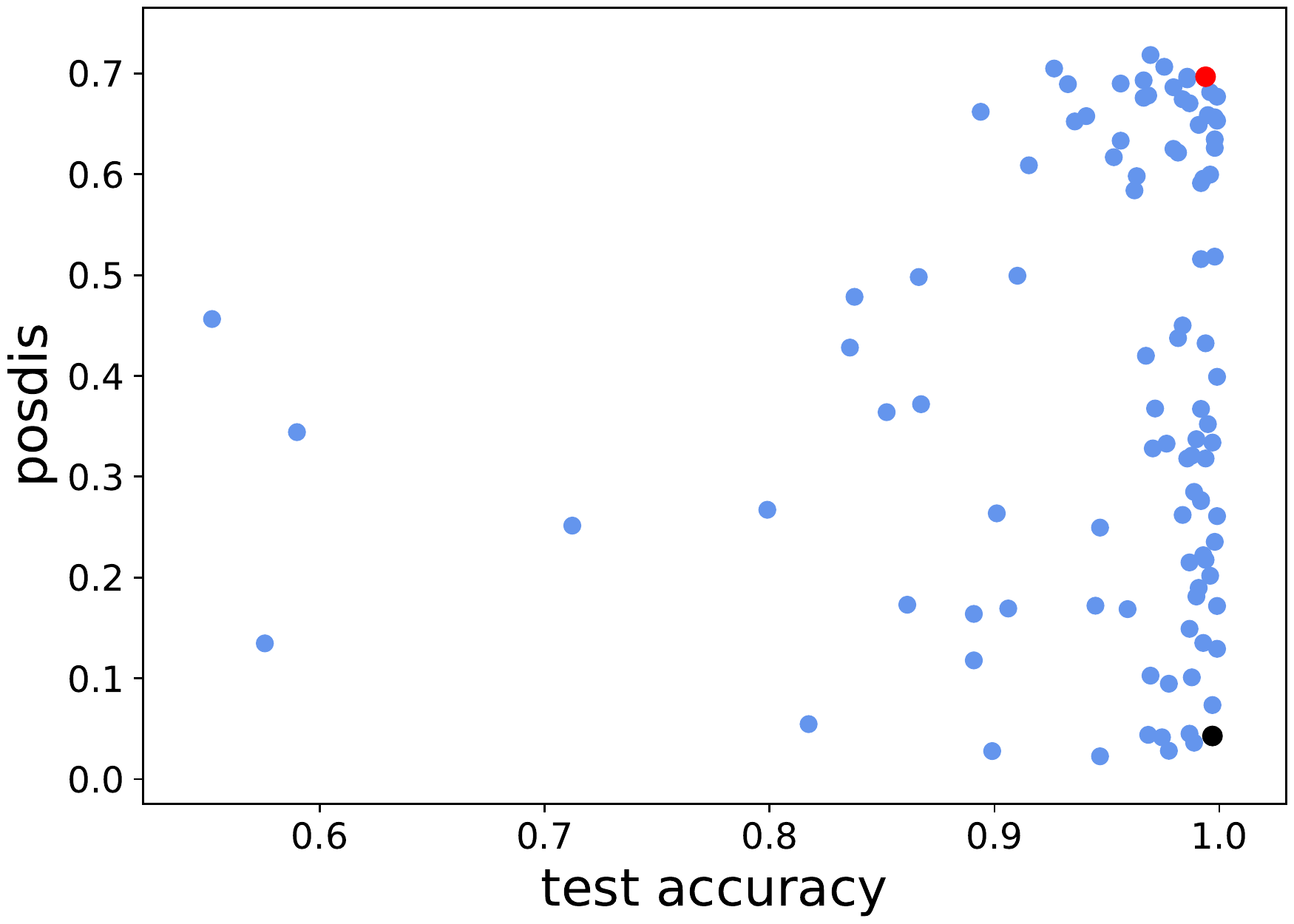}    
}
\subfigure[\textit{bosdis}]{
    \includegraphics[width=0.3\textwidth, keepaspectratio]{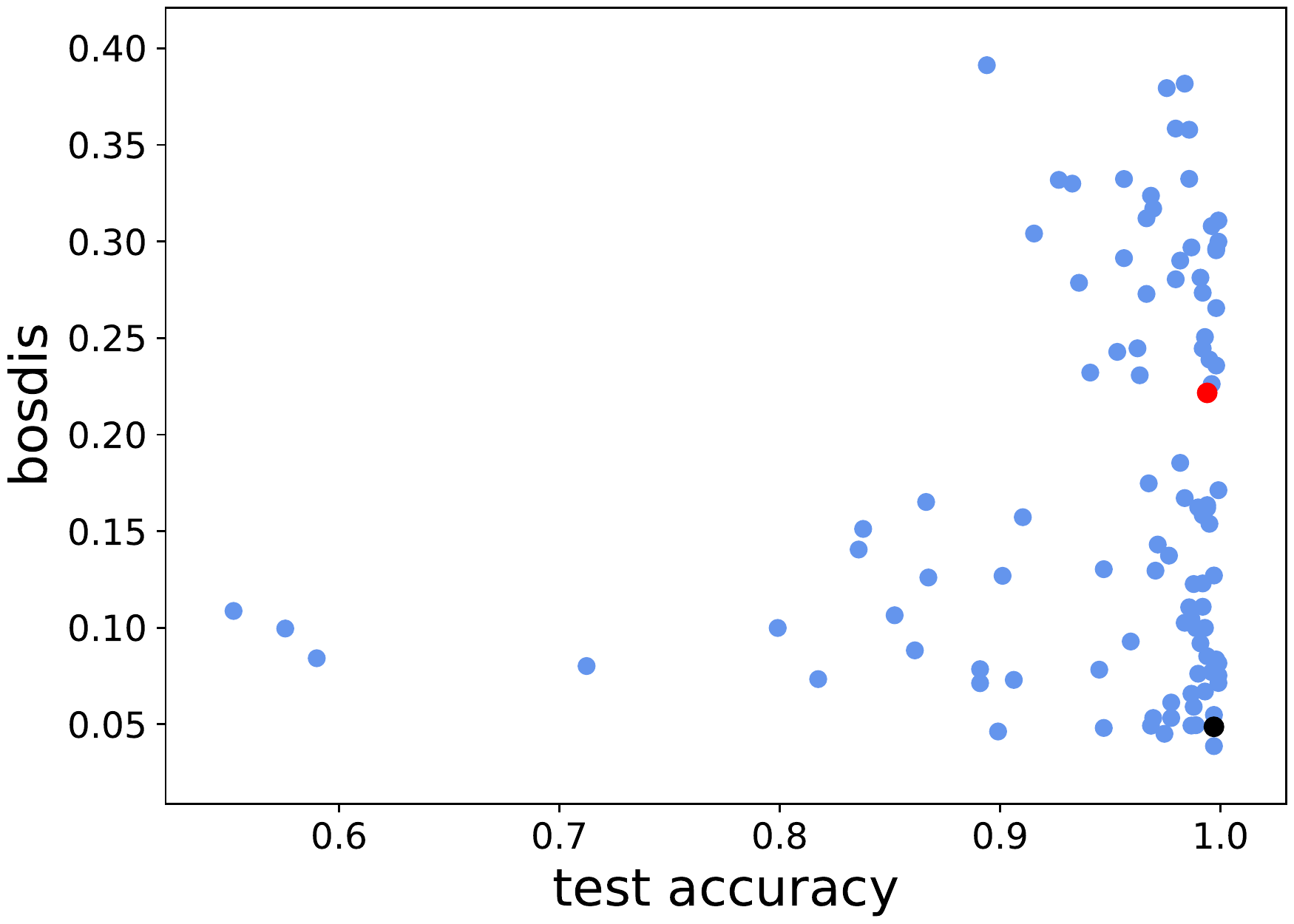}  
}
\subfigure[\textit{topsim}]{
    \includegraphics[width=0.3\textwidth, keepaspectratio]{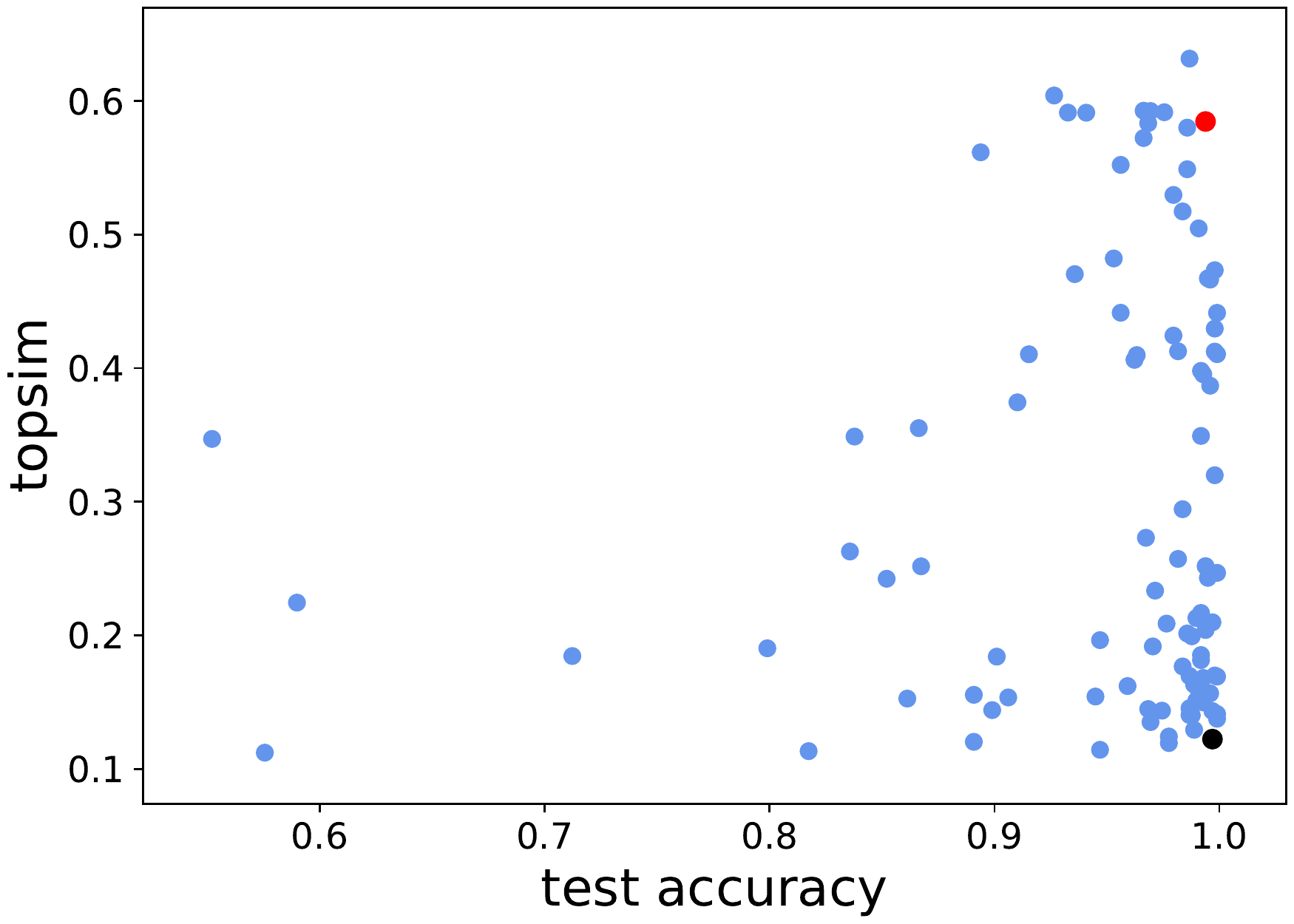}  
}
\caption{Compositionality in function of generalization. Each point represents a successful run in the ($i_{att}$=$2$, $i_{val}$=$100$, $c_{len}$=$3$, $c_{voc}$=$100$) setting. Red and black points  correspond respectively to the medium- and low-disentanglement languages analyzed in Section \ref{sGeneralizationCompositionality} and Appendix \ref{sec:A-looking-at-languages}.
  \label{fig:distributions}}
\end{figure*}

We further analyze  the  ($i_{att}$=$2$, $i_{val}$=$100$, $c_{len}$=$3$, $c_{voc}$=$100$) setting, as it has a large number of generalizing runs, and it is representative of the  general absence of correlation we also observe elsewhere. %
Fig.~\ref{fig:distributions} confirms that even non-compositional languages (w.r.t.~any definition of compositionality) can generalize well. Indeed, for very high test accuracy ($>98\%$), we witness a large spread of \textit{posdis} (between $0.02$ and $0.72$), \textit{bosdis} (between $0.03$ and $0.4$) and \textit{topsim} (between $0.11$ and $0.64$). In other words, deep agents are able to communicate about new attribute combinations while using non-compositional languages.
We note moreover that even the most compositional languages according to any metric are far from the theoretical maximum ($=1$ for all metrics).

We observe however that the top-left quadrants of Fig.~\ref{fig:distributions} panels are empty. In other words, it never happens that a highly compositional language has low accuracy. To verify this more thoroughly, for each compositionality measure $\mu$, we select those languages, among \emph{all converging runs in all configurations}, that have $\mu>0.5$, and compute the proportion of them that reaches high test accuracy ($>0.80$). We find that this ratio equates $0.90$, $0.50$, and $0.11$ for \textit{posdis}, \textit{bosdis}, and \textit{topsim} respectively. That is, while compositionality is not a necessary condition for generalization, it appears that the strongest form of compositionality, namely \emph{posdis}, is at least sufficient for generalization. This provides some evidence that  compositionality is still a desirable feature, as further discussed in Section~\ref{sBenefits}.

We gain further insights on what it means to generalize without full compositionality by taking a deeper look at the language shown
in red in Fig.~\ref{fig:distributions}, that has near-perfect
generalization accuracy ($>$99\%), and whose \textit{posdis} score
(0.70), while near the relative best, is still far from the theoretical
maximum (we focus on \emph{posdis} since it is the easiest
compositional strategy to qualitatively characterize). As its behavior
is partially interpretable, this ``medium-\emph{posdis}'' language
offered us clearer insights than more strongly entangled cases. We
partially analyze one of the latter in Appendix
\ref{sec:A-looking-at-languages}.

Note that, with ($i_{att}$=$2$, $i_{val}$=$100$), a ($c_{len}$=$2$,
$c_{voc}$=$100$) channel should suffice for a perfectly positionally
disentangled strategy. Why does the analyzed language use
($c_{len}$=$3$) instead? Looking at its mutual information profile (Appendix Table
\ref{tab:MI}), we observe that positions 2 and 3 (\emph{pos2} and
\emph{pos3}) are respectively denoting attributes 2 and 1 (\emph{att2}
and \emph{att1}): \emph{pos3} has high mutual information with \emph{att1} and low mutual information
with \emph{att2}; the opposite holds for \emph{pos2}.  The
remaining position, \emph{pos1}, could then be simply redundant with
respect to the others, or encode noise ignored by Receiver. %
However, this is not quite the case, as the language
settled instead for a form of ``leaky disentanglement''. The two
disentangled positions do most of the job, but the third, more
entangled one, is still necessary for perfect communication.

To see this, consider the ablations in Table
\ref{tab:shuffle-results_medium}. Look first at the \emph{top} block,
where the trained Receiver of the relevant run is fed messages
with the symbol in one original position preserved, the others
shuffled. Confirming that communication is largely happening by
disentangled means, preserving \emph{pos2} alone suffices to have
Receiver guessing a large majority of \emph{att2} values, and keeping
\emph{pos3} unchanged is enough to guess almost 90\% of \emph{att1}
values correctly. Conversely, preserving \emph{pos1} alone causes a
complete drop in accuracy for both attributes. However, neither
\emph{pos2} nor \emph{pos3} are sufficient on their own to perfectly
predict the corresponding attributes. Indeed, the results in the
\emph{bottom} block of the table (one symbol shuffled while the others stay in
their original position) confirm that \emph{pos1} carries useful
complementary information: when fixing the latter \emph{and either one of the other
positions}, we achieve 100\% accuracy for the relevant
attribute (\emph{att2} for \emph{pos1}+\emph{pos2} and \emph{att1} for
\emph{pos1}+\emph{pos3}), respectively.

In sum, \emph{pos2} and \emph{pos3} largely specialized as
predictors of \emph{att2} and \emph{att1}, respectively. However, they both have a
margin of ambiguity (in \emph{pos2} and \emph{pos3} there are 96 and
98 symbols effectively used, respectively, whereas a perfect 1-to-1
strategy would require 100). When the symbols in these positions do
not suffice, \emph{pos1}, that can refer to both attributes, serves a
disambiguating role. We quantified this complementary function as
follows. We define the cue validity of $s_p$ (symbol in position $p$)
w.r.t an attribute $a$ as $CV(s_p,a)=\max_{\bar a}{P(\bar a| s_p)}$,
where $\bar a$ iterates over all possible values of
$a$. $CV(s_{pos1},att2)$ is significantly higher in those (train/test)
messages where $CV(s_{pos2},att2)$ is below average. Similarly,
$CV(s_{pos1},att1)$ is significantly higher in messages where
$CV(s_{pos3},att1)$ is below average ($p\approx{}0$ in both cases). We
might add that, while there is a huge difference between our
simple emergent codes and natural languages, the latter are not
perfectly disentangled either, as they feature extensive
lexical ambiguity, typically resolved in a phrasal context
\cite{Piantadosi:etal:2012}.

\begin{table}[tbh]
  \centering
    \begin{tabular}{ll@{\,}|@{\,}r@{\,}|@{\,}r@{\,}|@{\,}r@{\,}}
      &           &\emph{att1}&\emph{att2}&\emph{both atts}\\
      \hline
      \emph{fixing}    &\emph{pos1}&1       &3       &0 \\
      \emph{1 position}&\emph{pos2}&1       &68      &0 \\
                       &\emph{pos3}&89      &1       &1 \\
      \hline
      \emph{shuffling} &\emph{pos1}&89     &69      &61\\     
      \emph{1 position}&\emph{pos2}&100    &3       &3\\     
                       &\emph{pos3}&1      &100     &1 \\
  \end{tabular}
  \caption{Feeding shuffled messages from the analyzed language to the
    corresponding trained Receiver. Average percentage accuracy across
    10 random shufflings (s.d. always $\approx{}0$)
    when: \emph{top}: symbols in all positions but one are shuffled
    across the data-set; \emph{bottom}: symbols in a single position
    are shuffled across the data-set. The data-set includes all
    training and test messages produced by the trained Sender and
    correctly decoded in their original form by Receiver ($>$99\% of
    total messages).}
  \label{tab:shuffle-results_medium}
\end{table}

\section{Compositionality and ease of transmission}
\label{sBenefits}
The need to generalize to new composite inputs does not appear to constitute a
sufficient pressure to develop a compositional language. Given that
compositionality is ubiquitous in natural language, we conjecture that it has other beneficial properties, making it
advantageous once agents chanced upon it. Compositional codes are
certainly easier to read out by humans (as shown by our own difficulty
in qualitatively analyzing highly entangled languages), and we might
hypothesize that this ease-of-decoding is shared by
computational agents. A long tradition of subject studies and
computational simulations has shown that the need to transmit a
language across multiple generations or to populations of new learners
results in the language being more compositional
\cite[e.g.,][]{kirby2001spontaneous,Kirby:etal:2015,Verhoef:etal:2015,Cornish:etal:2017,cogswell2019,Guo:etal:2019,Li2019}. Our
next experiments are closely related to this earlier work, but we
adopt the opposite perspective. Instead of asking whether the pressure
to transmit a language will make it more compositional, we test
whether languages that have already emerged as compositional, being easier to decode, are more readily transmitted to new learners.\footnote{\citet{Li2019}
  established this for hand-crafted languages; we extend the
  result to spontaneously emerging ones.}

Specifically, we run $30$ games in the largest input setting ($i_{att}$=$2$, $i_{val}$=$100$), varying the channel parameters. We select the pairs of agents that achieved a high level of generalization accuracy ($\ge$0.80). Next, following the paradigm of \citet{Li2019}, we freeze Sender, and train a new Receiver from scratch. We repeat this process $3$ times per game, initializing new Receivers with different random seeds. Once the newly formed pair of agents is successful on the training set, we measure its test accuracy. We also report speed of learning, measured by area under the epochs vs.\ training accuracy curve.  We experiment with three Receiver architectures. The first two, GRU (500) and GRU (50), are GRUs with hidden layer sizes of 500 (identical to the original Receiver) and 50, respectively. The third is a two-layer Feed-Forward Network (FFN) with a ReLu non-linearity and hidden size 500. The latter Receiver takes the flattened one-hot representation of the message as its input. This setup allows probing ease of language transmission across models of different complexity. We leave the study of language propagation across multiple generations of speakers to future work.


\begin{table*}[t]
\centering
\setlength{\tabcolsep}{3.5pt}
\begin{tabular}{lcccccccccccc}
\toprule
 & \multicolumn{3}{c}{\textit{posdis}} && \multicolumn{3}{c}{\textit{bosdis}} && \multicolumn{3}{c}{\textit{topsim}}\\
\cmidrule{2-4} \cmidrule{6-8} \cmidrule{10-12}
& {GRU}\small{(500)} & {GRU}\small{(50)} & {FFN} && {GRU}\small{(500)} & {GRU}\small{(50)} & {FFN} && {GRU}\small{(500)} & {GRU}\small{(50)} & {FFN} \\
\midrule
\small{Learning Speed} & 0.87 & 0.71 & 0.35 && 0.85& 0.68 & 0.33 && 0.87 & 0.71& 0.35  \\
\small{Generalization} & 0.80 & 0.55 & 0.50 && 0.81 & 0.55 & 0.51 && 0.79 & 0.54& 0.48  \\
\bottomrule
\end{tabular}
\caption{Spearman correlation between compositionality metrics and ease-of-transmission measures for ($i_{att}$=$2$, $i_{val}$=$100$, $c_{len}$=$3$, $c_{voc}$=$100$). All values are statistically significant ($p < 0.01$). \label{tab:retrain}}
\end{table*}
Results in the same setting studied in Section~\ref{sGeneralizationCompositionality} are presented in Table~\ref{tab:retrain} (experiments with other setups are in Appendix \ref{sec:retrain}). Both learning speed and generalization accuracy of new Receivers are \textit{strongly positively correlated with degree of compositionality}. The observed correlations reach values almost as high as $0.90$ for learning speed and $0.80$ for generalization, supporting our hypothesis that, when emergent languages are compositional, they are simpler to understand for new agents, including smaller ones (GRU (50)), and those with a different architecture (FFN). 

\section{Discussion}

\paragraph{The natural emergence of generalization} There has been
much discussion on the generalization capabilities of neural networks,
particularly in linguistic tasks where humans rely on compositionality
\cite[e.g.,][]{Fodor:Lepore:2002,Marcus:2003,vanderVelde:etal:2004,Brakel:Frank:2009,Kottur:etal:2017,Lake:Baroni:2017,Andreas:2019,Hupkes:etal:2019,Resnick2019}. In
our setting, the emergence of generalization is very strongly
correlated with variety of the input environment. While this
result should be replicated in different conditions, it suggests that
it is dangerous to study the generalization abilities of neural
networks in ``thought experiment'' setups where they are only
exposed to a small pool of carefully-crafted examples. Before
concluding that garden-variety neural networks do not generalize, the simple strategy
of exposing them to a richer input should always be
tried. Indeed, even studies of the origin of human language conjecture
that the latter did not develop sophisticated generalization
mechanisms until pressures from an increasingly complex environment
forced it to evolve in that direction
\cite{Bickerton:2014,Hurford:2014}.

\paragraph{Generalization without compositionality} Our most important
result is that \emph{there is virtually no correlation} between
whether emergent languages are able to generalize to novel composite
inputs and the presence of compositionality in their messages
(\citet{Andreas:2019} noted in passing the emergence of
non-compositional generalizing languages, but did not explore this
phenomenon systematically). Supporting generalization to new composite
inputs is seen as one of the core purposes of compositionality in
natural language \cite[e.g.,][]{Pagin:Westerstahl:2010}. While there
is no doubt that compositional languages do support generalization, we
also found other systems spontaneously arising that generalize without
being compositional, at least according to our intuitive measures of
compositionality. This has implications for the ongoing debate on the
origins of compositionality in natural language, \cite[e.g.,][and
references there]{Townsend:etal:2018}, as it suggests that the need to
generalize alone might not constitute a sufficient pressure to develop
a fully compositional language. Our result might also speak to those
linguists who are exploring the non-fully-compositional corners
of natural language \cite[e.g.,][]{Goldberg:2019}. A thorough
investigation of neural network codes that can generalize while
being partially entangled might shed light on similar phenomena in
human languages. Finally, and perhaps most importantly, recent
interest in compositionality among AI researchers stems from the assumption
that compositionality is crucial to achieve good generalization through language
\cite[e.g.,][]{Lake:Baroni:2017,Lazaridou:etal:2018,Baan:etal:2019}. Our
results suggest that the pursuit of generalization might be separated
from that of compositionality, a point also recently made by
\citet{Kharitonov:Baroni:2020} through hand-crafted simulations.

\paragraph{What is compositionality good for?} We observed that
positional disentanglement, while not necessary, is sufficient for
generalization. If agents develop a compositional language, they are
then very likely to be able to use it correctly to refer to novel
inputs. This supports the intuition that compositional languages are
easier to fully understand. Indeed, when training new agents on
emerged languages that generalize, it is much more likely that the new
agents will learn them fast and thoroughly (i.e., they will be able to
understand expressions referring to novel inputs) if the languages are
already compositional according to our measures. That language
transmission increases pressure for structured representations is an
established fact
\cite[e.g.,][]{Kirby:etal:2015,Cornish:etal:2017}. Here, we reversed
the arrow of causality and showed that, if compositionality emerges
(due to chance during initial language development), it will make a
language easier to transmit to new agents. Compositionality might act
like a ``dominant'' genetic feature: it might arise by a random
mutation but, once present, it will survive and thrive, as it
guarantees that languages possessing it will generalize and will be
easier to learn. From an AI perspective, this suggests that trying to
enforce compositionality during language emergence will increase the
odds of developing languages that are quickly usable by wide
communities of artificial agents, that might be endowed with different
architectures. From the linguistic perspective, our results suggest an
alternative view of the relation between compositionality and language
transmission--one in which the former might arise by chance or due to
other factors, but then makes the resulting language much easier to be
spread. 

\paragraph{Compositionality and disentanglement} Language is a way to
\emph{represent} meaning through discrete symbols. It is thus worth
exploring the link between the area of language emergence and that of
representation learning \citep{Bengio2013}. 
We took this route, borrowing ideas
from 
research on disentangled representations to craft our compositionality
measures. We focused in particular on the intuition that, if
emergent languages must denote ensembles of primitive input elements,
they are compositional when they use symbols to univocally denote
input elements independently of each other.

While the new measures we proposed are not highly correlated with
topographic similarity, in most of our experiments they did not behave
significantly differently from the latter. On the one hand, given that
topographic similarity is an established way to quantify
compositionality, this serves as a sanity check on the new
measures. On the other, we are disappointed that we did not find
more significant differences between the three measures.

Interestingly one of the ways in which they did differ is
that, when a language is positionally disentangled, (and, to a lesser
extent, bag-of-symbols disentangled), it is very likely that the
language will be able to generalize--a guarantee we don't have from
less informative 
topographic similarity.

The representation learning literature is not only proposing
disentanglement measures, but also ways to encourage emergence of
disentanglement in learned representations.  As we argued that
compositionality has, after all, desirable properties, future work
could adapt methods for learning disentangled
representations~\cite[e.g.,][]{Higgins2017,Kim2018} to let (more)
compositional languages emerge.


\section*{Acknowledgments}

We thank the reviewers for feedback that helped us to make the paper clearer.

\bibliography{marco,other}
\bibliographystyle{acl_natbib}

%
%

\section{Supplementary material}

\subsection{Grid search over ($i_{att}$, $i_{val}$, $c_{len}$, $c_{voc}$)}
\label{sec:gridSearch}
We report in Table~\ref{tab:dimensions} the different ($i_{att}$, $i_{val}$, $c_{len}$, $c_{voc}$) combinations we explored. They were picked according to the following criteria:
\begin{itemize}
    \item $|C| \geq |I|$ so that agents are endowed with enough different messages to refer to all inputs;
    \item discard some $|C|>>|I|$ so that we have approximately the same number of settings per ($i_{att}$, $i_{val}$) (between $13$ and $15$ different ($c_{voc}$, $c_{len}$));
    \item include some ($c_{voc}$, $c_{len}$) that are large enough that they can be tested with all the considered ($i_{att}$, $i_{val}$).

\end{itemize}

Unless it is mentioned explicitly, we run $10$ different initializations per setting.

Table~\ref{tab:dimensions} shows that, for large $|I|$, GRU-agents need $|C|$ \emph{strictly} larger than $|I|$. 
This suggests that, for large $|I|$, the emergence of a perfectly
non-ambiguous compositional languages, where each message symbol
denotes only one attribute value \emph{and} each value attribute is
denoted by only one message symbol, is impossible.

\begin{table*}[t]
\centering
\setlength{\tabcolsep}{4pt}
\begin{tabular}{ll|cccccccccccccccccccc}
\toprule
 \multicolumn{2}{c|}{\multirow{2}{*}{\diagbox[width=8em, height=3em]{\hspace{1.5em}\small{($i_{val}$, $i_{att}$)}}{$c_{voc}$\vspace{.1em}\\$c_{len}$}}} &
& \multicolumn{4}{c}{5} && \multicolumn{4}{c}{10} && \multicolumn{4}{c}{50} && \multicolumn{4}{c}{100}\\
\cmidrule{3-6} \cmidrule{8-11} \cmidrule{13-16} \cmidrule{18-21}
&& $2$ &$3$&$4$&$\{6$,$8\}$&&$2$ &$3$&$4$&$\{6$,$8\}$ &&$2$ &$3$&$4$&$\{6$,$8\}$ &&$2$ & $3$ & $4$ & $\{6$,$8\}$ \\
\midrule
\multicolumn{2}{c|}{($4$,$4$)} &&  & X& X&&  &X &X& X&& X& & & X && X& & & X \\
\multicolumn{2}{c|}{($5$,$2$)} &X& X & X& X&& X &X &X& X&& X& &  &X && X& & & X& \\
\multicolumn{2}{c|}{($5$,$3$)}  && X & X& X&&  &X &X& X&& X& &  &X && X& & & X& \\
\multicolumn{2}{c|}{($5$,$4$)} &&  & X& X&&  &X &X& X&& X& &  &X && X& & & X& \\
\multicolumn{2}{c|}{($10$,$2$)} && - & X& X&& X &X &X& X&& X& &  &X && X& & & X& \\
\multicolumn{2}{c|}{($10$,$3$)} &&  & & X&&  &- &X& X&& X& X& X &X && X& & & X& \\
\multicolumn{2}{c|}{($10$,$4$)} &&  & & \{-, X\}&&  & &-& X&& & X& X &X && -& X& X& X& \\
\multicolumn{2}{c|}{($16$,$2$)} &&  & -& X&&  &X &X& X&& X& & & X && X& & & X \\
\multicolumn{2}{c|}{($25$,$2$)}&&  & -& X&&  &- &X& X&& X& &  &X && X& & & X& \\
\multicolumn{2}{c|}{($50$,$2$)} &&  & & X&&  & &-& X&& -& X& X &X && X& X& X& X& \\
\multicolumn{2}{c|}{($100$,$2$)}&&  & & \{-, X\}&&  & &-& X&& & X& X &X && -& X& X& X& \\
\bottomrule
\end{tabular}
\caption{Grid search. `X' indicates tested settings with at least one successful run. `-' indicates tested settings without any successful run. Finally, blank cells correspond to settings that were not explored for the reasons indicated in the text.  \label{tab:dimensions}}
\end{table*}

\subsection{Behavior of the  compositionality measures on hand-crafted miniature languages}
\label{sec:minLang}
We construct $3$ simple miniature languages to illustrate the different behaviors of \textit{topsim}, \textit{posdis} and \textit{bosdis}: Lang1, Lang2 and Lang3. We fix $i_{att}=2$, $i_{val}=4$, $c_{len}=3$ and $c_{voc}=8$.\footnote{Only Lang3 uses the whole available $c_{voc}$} Table \ref{tab:minLang} shows the input-message mappings of each language and reports their degree of compositionality. Note that  all languages respect a bijective mapping between inputs and messages.

Lang1 is perfectly \textit{posdis}-compositional
(\textit{posdis}=1). However, \textit{topsim }$<1$, as $2$ symbols
encode one attribute (we need the first two symbols to recover the
value of the first attribute). Lang1 is penalized by \textit{topsim}
because it does not have a one-to-one attribute-position mapping; a
feature that arguably is orthogonal to compositionality.

Lang2 and Lang3 are equally \textit{topsim}-compositional. Nonetheless, they differ fundamentally in terms of the type of compositionality they feature. If Lang2 is more \textit{posdis}-compositional, Lang3 is perfectly \textit{bosdis}-compositional.

\begin{table}
\centering
\begin{tabular}{lccccccc}
\toprule
Input & Lang1 & Lang2 & Lang3\\
\midrule
    0,0 & 0,0,0 & 0,0,0 & 0,0,4\\ 
    0,1 & 0,0,1 & 0,0,1 & 0,0,5\\ 
    0,2 & 0,0,2 & 0,0,2 & 0,0,6\\ 
    0,3 & 0,0,3 & 0,0,3 & 0,0,7\\ 
    1,0 & 0,1,0 & 1,2,0 & 1,4,1\\ 
    1,1 & 0,1,1 & 1,2,1 & 1,5,1\\ 
    1,2 & 0,1,2 & 1,2,2 & 1,6,1\\ 
    1,3 & 0,1,3 & 1,2,3 & 1,7,1\\ 
    2,0 & 2,0,0 & 2,3,0 & 2,4,2\\ 
    2,1 & 2,0,1 & 2,3,1 & 2,5,2\\ 
    2,2 & 2,0,2 & 2,3,2 & 2,6,2\\ 
    2,3 & 2,0,3 & 2,3,3 & 2,7,2\\ 
    3,0 & 2,1,0 & 3,1,0 & 3,4,3\\ 
    3,1 & 2,1,1 & 3,1,1 & 3,3,5\\ 
    3,2 & 2,1,2 & 3,2,1 & 3,3,6\\ 
    3,3 & 2,1,3 & 3,3,1 & 3,3,7\\ 
\hline
\hline
    \textit{topsim} & $0.82$ & $0.75$ & $0.75$\\
    \textit{posdis} & $1$ & $0.79$ &  $0.43$\\
    \textit{bosdis} & $0.42$ & $0.13$ & $1$ \\
\bottomrule
\end{tabular}
\caption{Input-message mappings and compositionality measures for the miniature languages.\label{tab:minLang}}
\end{table}

\subsection{Generalization for different agents' capacity}
\label{sec:robust}
\begin{figure}
   \centering
      \includegraphics[width=\columnwidth]{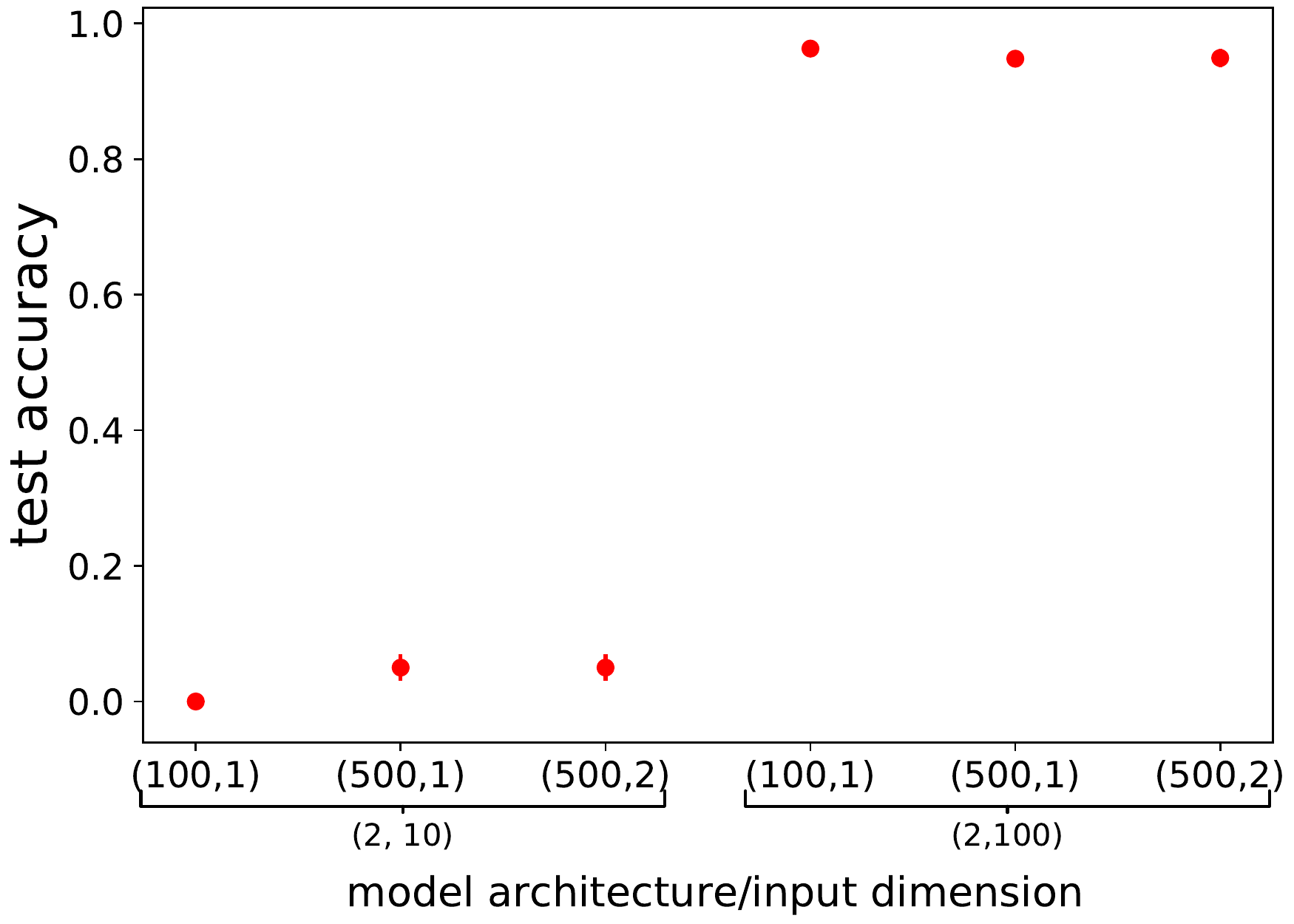}
      \caption{Average accuracy on unseen combinations as a function of agents capacity ((hidden size, number of layers)) for input sizes ($i_{att}=2$, $i_{val}=10$) and ($i_{att}=2$, $i_{val}=100$). Vertical bars represent SEM.}
\label{fig:generalizationarchi}
 \end{figure}

We demonstrated in the main paper that agent's generalization correlates with input size. In fact, agents can successfully reconstruct new attribute combinations if trained on large input spaces. This could be due to agents overfitting when presented with few training samples. To test this hypothesis, we repeat the training/evaluation experiments with GRU agents of different capacities in the following settings: ($i_{att}$=$2$, $i_{val}$=$10$), a small input space where agents do not generalize; and ($i_{att}$=$2$, $i_{val}$=$100$), a large input space where agents generalize.\footnote{We only report experiments with GRUs, but the same results were replicated with differently-sized LSTMs.} Fig.~\ref{fig:generalizationarchi} shows that, even for small-capacity agents (one-layer GRU with hidden state of size $100$), test accuracy is 0 for ($i_{att}$=$2$, $i_{val}$=$10$). Moreover, agents do not overfit when trained on ($i_{att}$=$2$, $i_{val}$=$100$) even with two-layer GRUs with hidden state of size $500$.

\subsection{Input space density}
\label{sec:density}
We showed in the main paper that generalization positively correlates
with $|I|$. We further investigate here whether it is simply the
increasing absolute number of distinct training samples that is at the
root of this phenomenon, or whether the variety of seen inputs also
plays a role, independently of absolute input size.

To verify this, we design an experiment where we keep the absolute
number of distinct input samples constant, but we change their
\emph{density}, defined as the proportion of sampled items over the the size of the space they are
sampled from. When sampling points from a small space, on average
each value of an attribute will occur with a larger range of values
from other attributes, compared to a larger space, which might provide more evidence about the
combinatorial nature of the underlying space.

\begin{figure*}[ht]
\centering
\subfigure[density=$1$]{
    \includegraphics[width=0.3\textwidth, keepaspectratio]{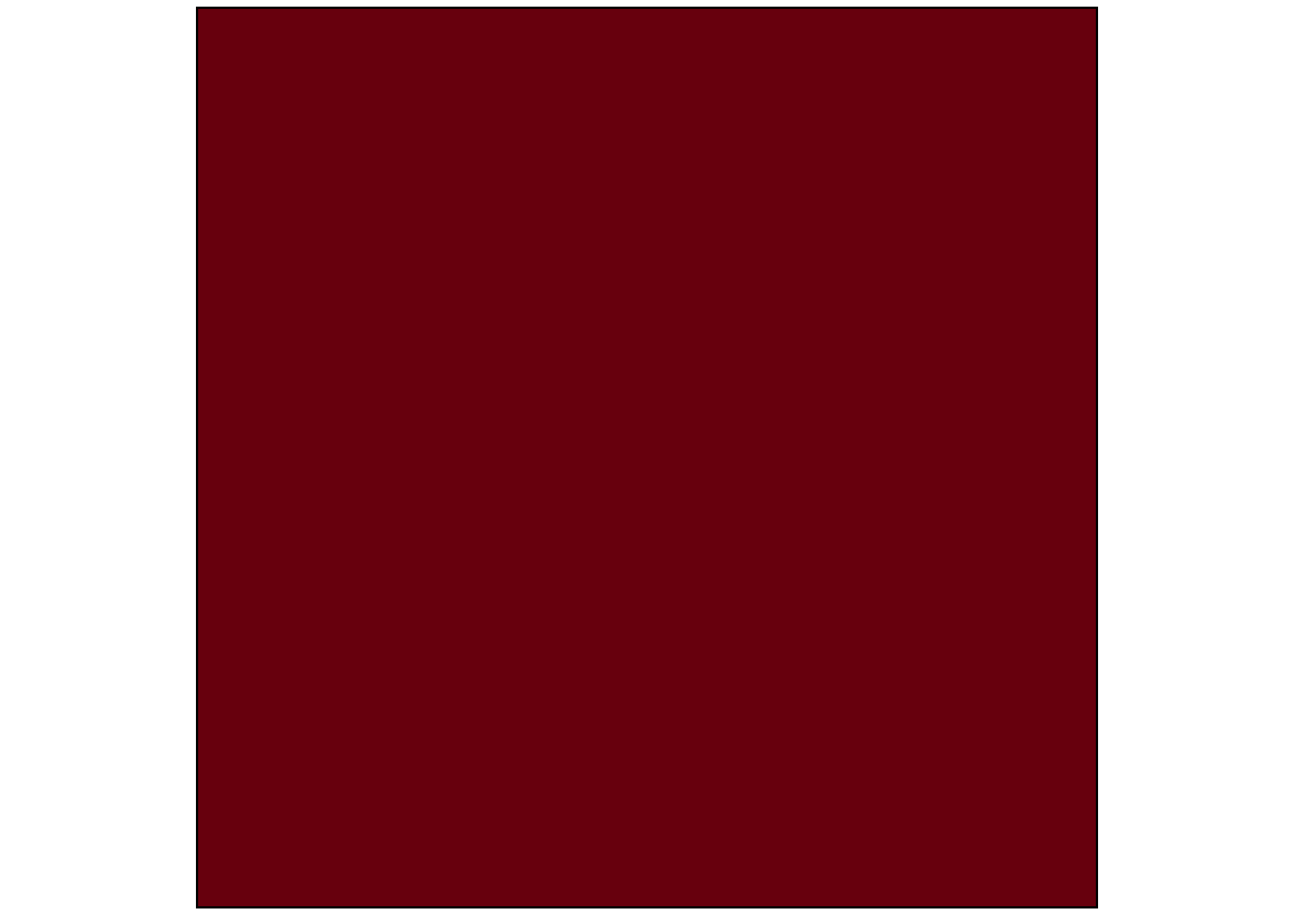}    
}
\subfigure[density=$0.51$]{
    \includegraphics[width=0.3\textwidth, keepaspectratio]{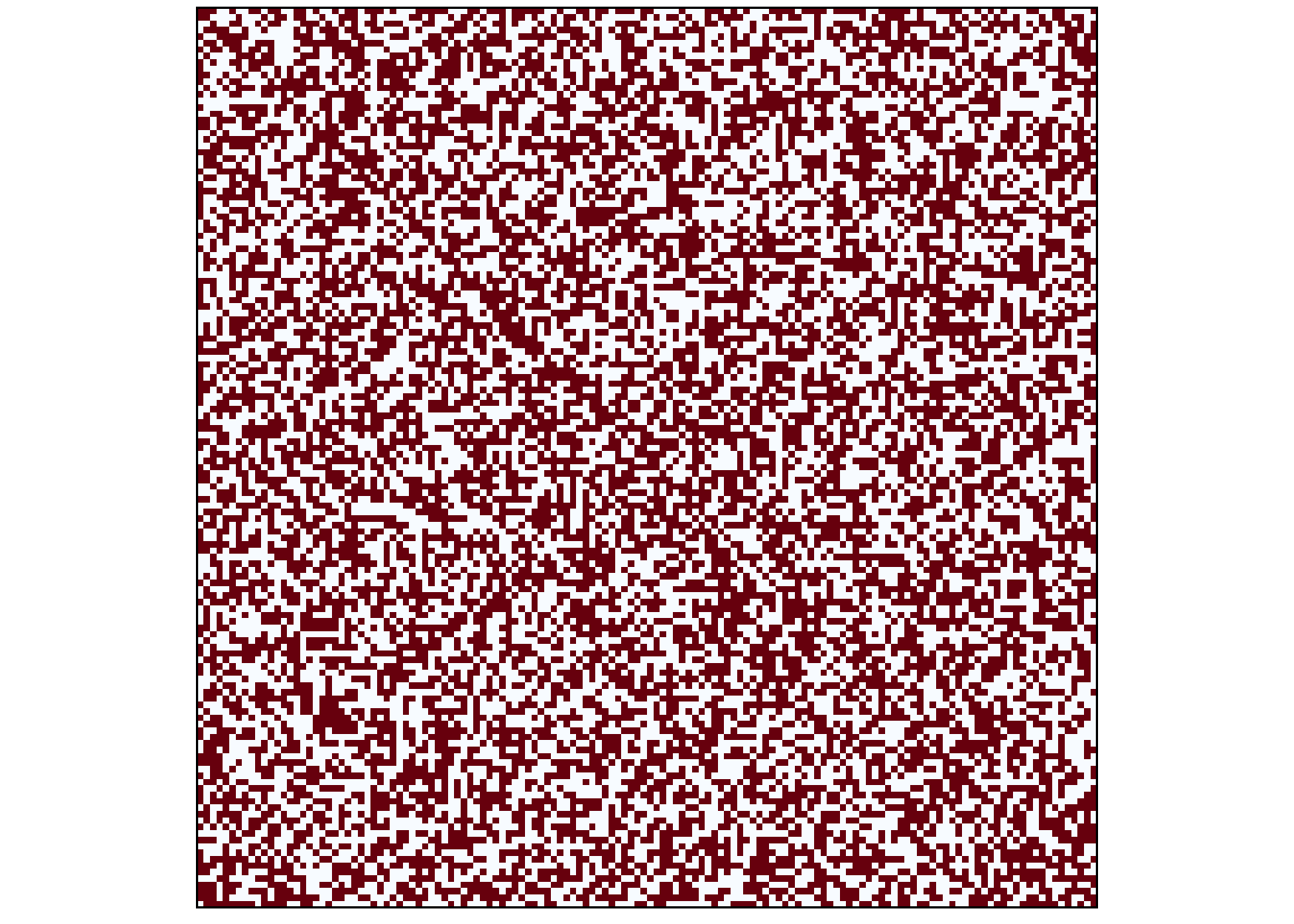}  
}
\subfigure[density=$0.25$]{
    \includegraphics[width=0.3\textwidth, keepaspectratio]{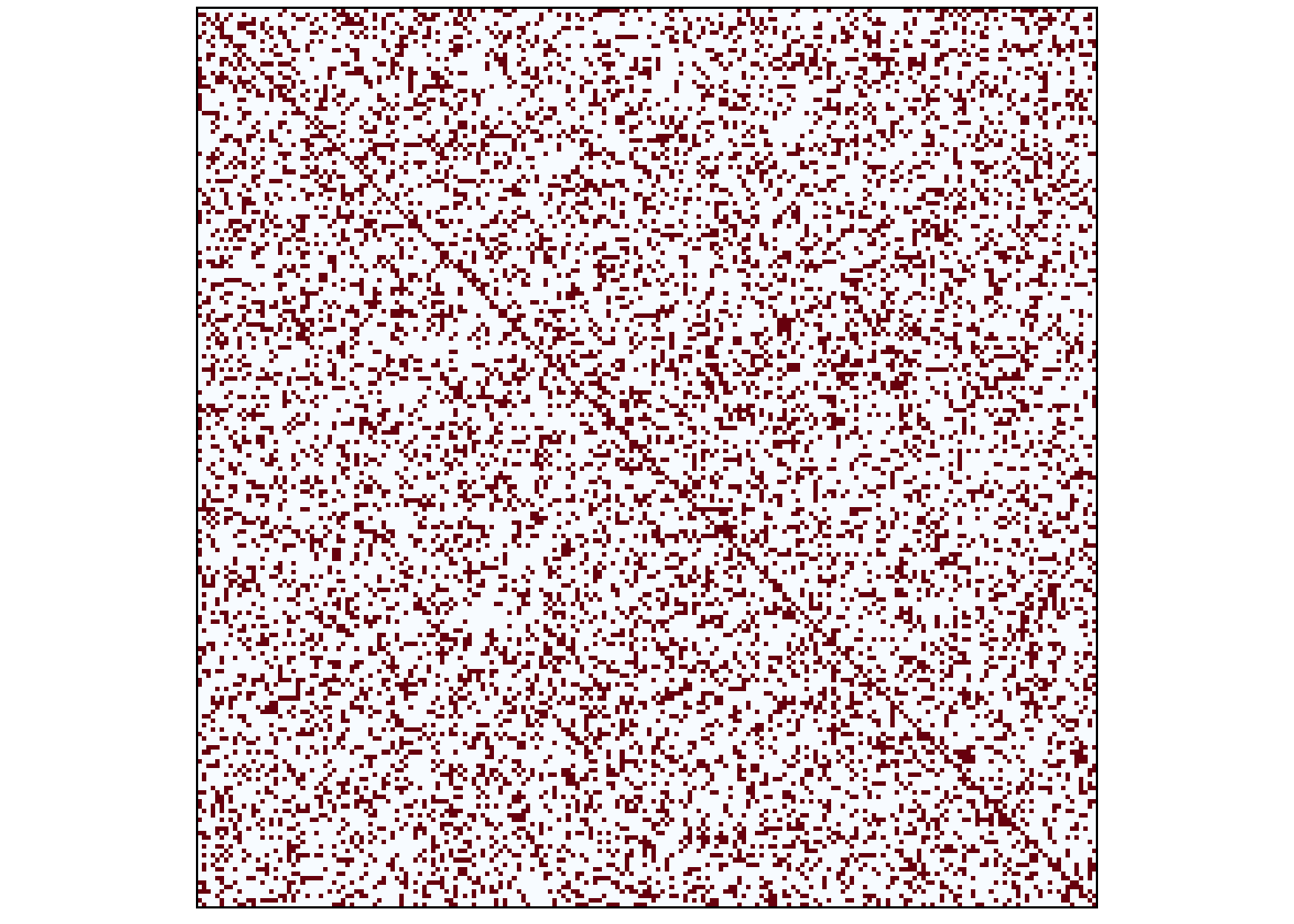}  
}
\caption{Sampling the same number of input instances ($=10000$) with
  different densities. The axes of the shown matrices represent the
  values of two attributes, with the dark-red cells standing for
  inputs that were sampled. We ensure that each value of each
  attribute is picked at least once by always sampling the full
  diagonal.\label{fig:density}}
\end{figure*}

In practice, we fix ($c_{len}$=$3$, $c_{voc}$=$100$, $i_{att}$=$2$)
and sample $10000$ points from spaces with $i_{val}$=$100$
(density=$1$), $i_{val}$=$140$ (density=$0.51$) and $i_{val}$=$200$
(density=$0.25$), respectively. As usual, we use 90\% of the data for
training, 10\% for testing. In all cases, we make sure that all values
are seen at least once during training (as visually illustrated in
Fig.~\ref{fig:density}).

We obtain test accuracies of $92.7\%$, $66.7\%$ and $22.8\%$ for
densities $1$, $0.51$ and $0.25$ respectively. That is, the high
generalization observed in the main paper is (also) a consequence of
density, and hence combinatorial variety, of the inputs the agents are trained on, and not (only) of the
number of training examples.

\subsection{Impact of channel capacity on generalization}
\label{sec:channelgen}
We showed in the main paper that generalization is very sensitive to input size. In this section, we focus on the relation between channel capacity $|C|$ and generalization. 

 \begin{figure}
   \centering
      \includegraphics[width=0.8\columnwidth]{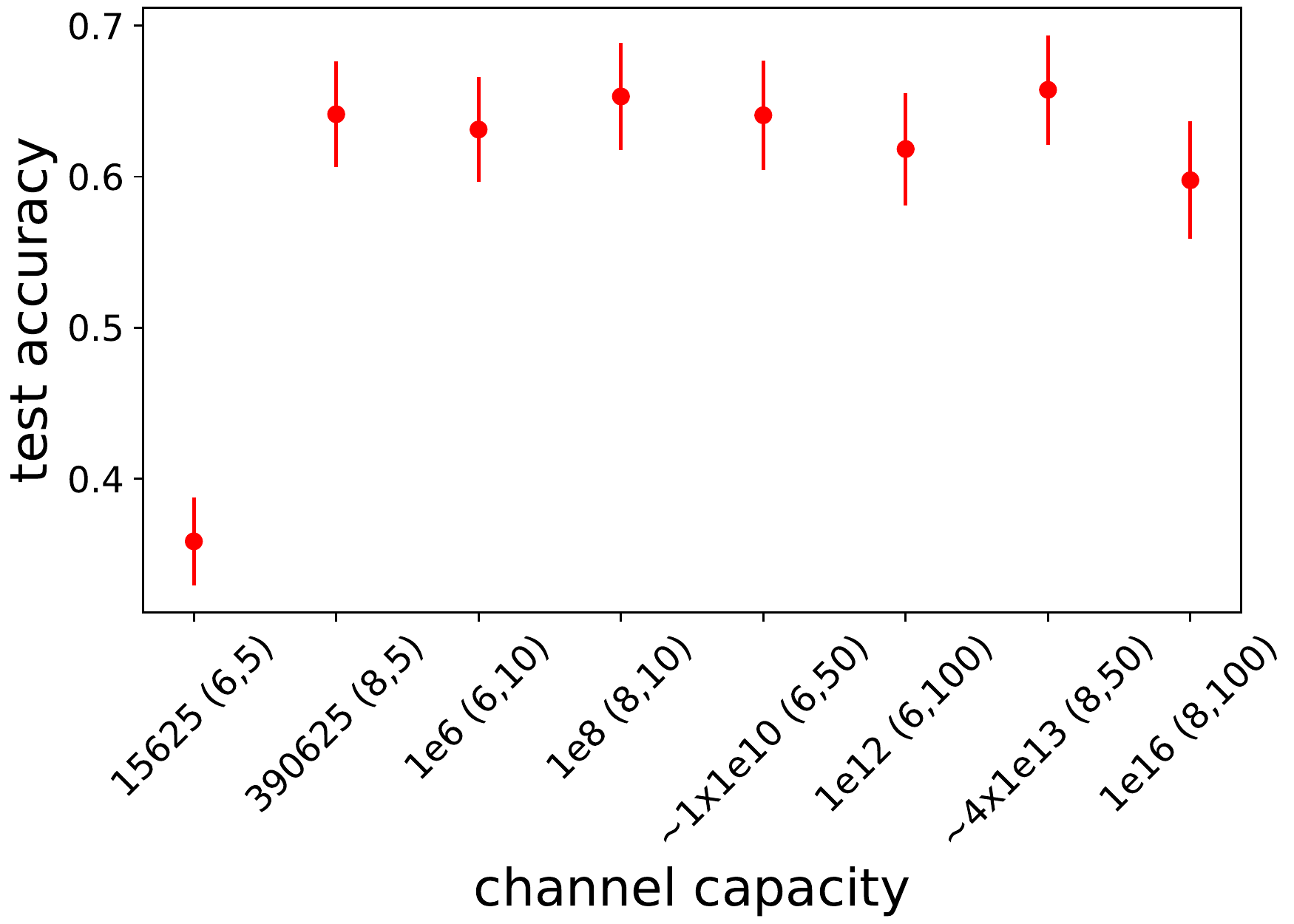}
      \caption{Average accuracy on unseen combinations as a function of channel capacity of the successful runs. The x-axis is ordered by increasing channel capacity. In the brackets we note ($c_{len}$, $c_{voc}$). 
      Vertical bars represent SEM.}
\label{fig:generalization2}
 \end{figure}

\begin{figure*}[ht]
\centering
\hspace{-2.2cm}
\subfigure[($i_{att}=2$, $i_{val}=50$)]{
    \hspace{-.4cm}
    \includegraphics[width=0.45\textwidth, keepaspectratio]{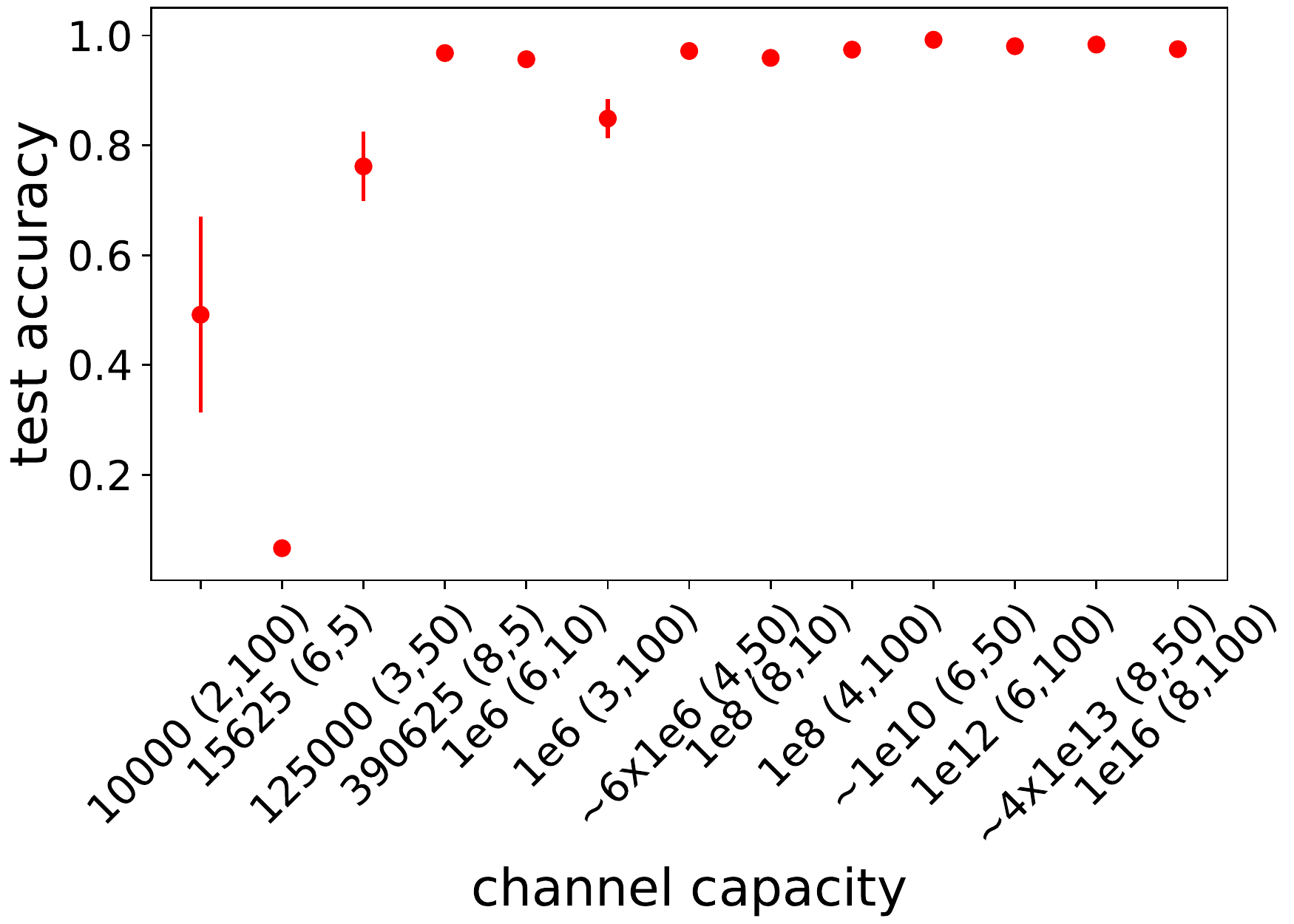}    
}
\subfigure[($i_{att}=2$, $i_{val}=100$)]{
    \hspace{-.4cm}
    \includegraphics[width=0.45\textwidth, keepaspectratio]{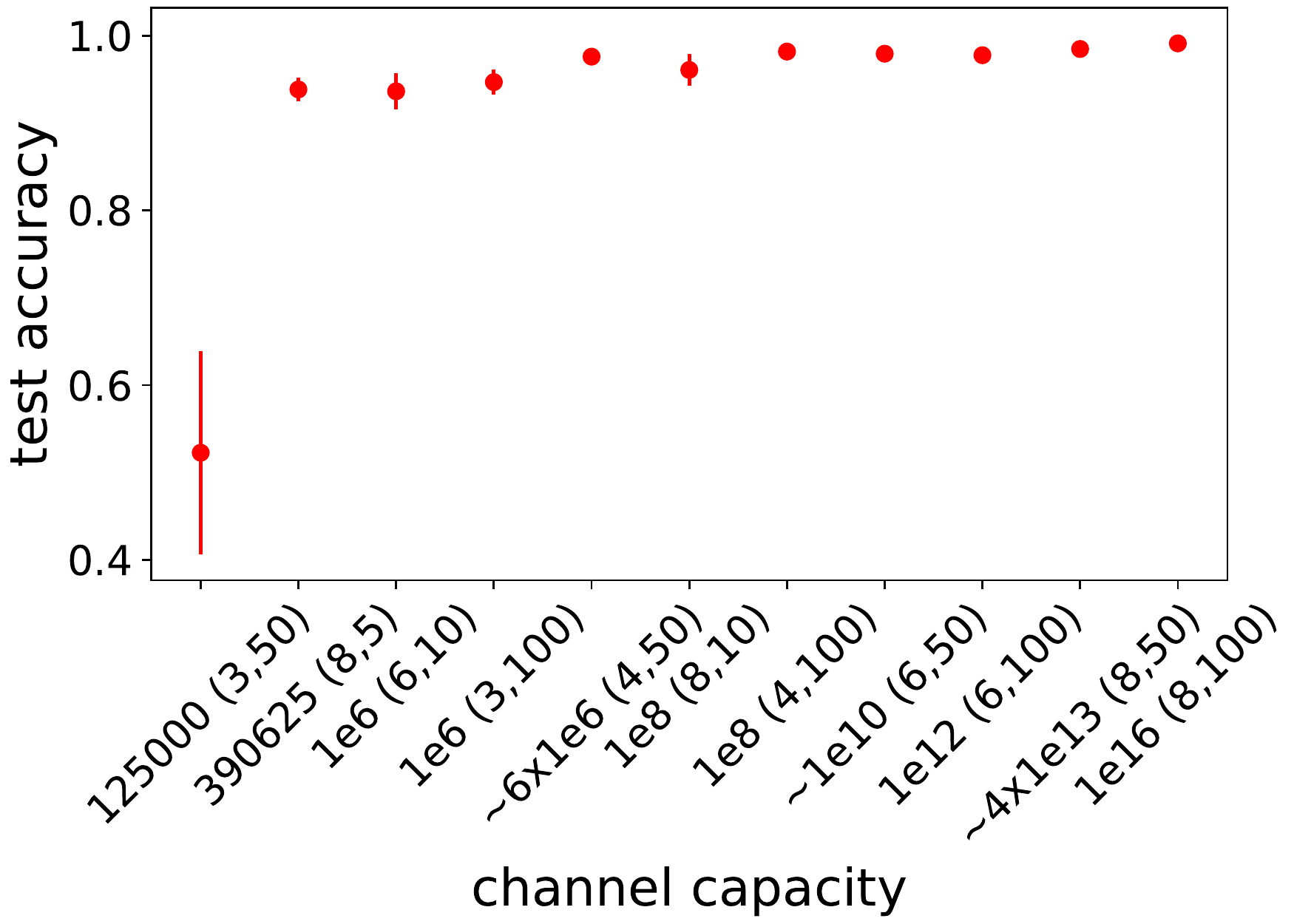} 
}
\hspace{-2cm}
\caption{Average accuracy on unseen combinations as a function of channel capacity of the successful runs for two different ($i_{att}$, $i_{val}$). The x-axis is ordered by increasing channel capacity. In the brackets we note ($c_{len}$, $c_{voc}$).
Vertical bars represent SEM. 
}\label{fig:generalization3}
\end{figure*}

First, when we aggregate across input sizes, Fig.~\ref{fig:generalization2} shows that $|C|$ has a just small effect on generalization, with a low Spearman correlation $\rho=0.14$. Next, if we study this relation for specific large $|I|$ (where we observe generalization), we notice in Fig.~\ref{fig:generalization3} that agents need to be endowed with a $|C|$ above a certain threshold, with $\frac{|C|}{|I|}>1$, in order to achieve almost perfect generalization. 
Moreover, contradicting previous claims \cite[e.g.,][]{Kottur:etal:2017}, having $|C|>>|I|$ does not harm generalization.


\subsection{Impact of channel capacity on the compositionality measures}
\label{sec:dependence}
\begin{figure}
   \centering
      \includegraphics[width=\columnwidth]{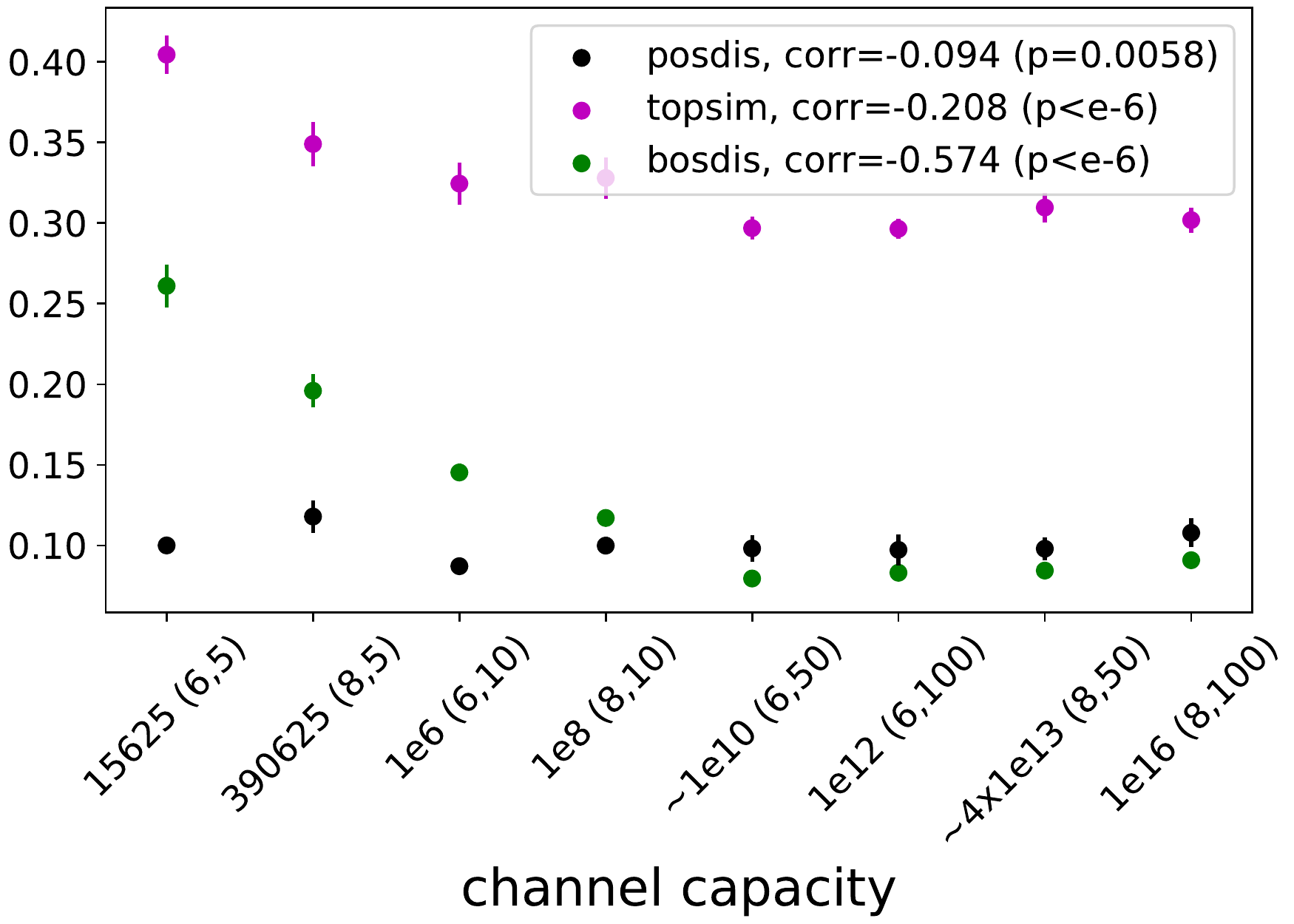}
      \caption{Average of different compositionality measures in function of channel capacity ($c_{voc}$, $c_{len}$)). Vertical bars represent SEM.}
\label{fig:measureschannel}
 \end{figure}

A good  compositionality measure should describe the structure of the language independently of the used channel, so the corresponding score should not be greatly affected by $|C|$. However, Fig.~\ref{fig:measureschannel} shows clear negative correlations of  both \textit{topsim} and \textit{bosdis} with $|C|$. 
 
\subsection{Analysis of example medium- and low-posdis languages}
\label{sec:A-looking-at-languages}

We present more data about the \emph{medium-posdis} language analyzed
in the main article, and we provide comparable evidence for a language
with similarly excellent generalization ($>$99\%) but very low posdis
(0.05), that we will call here \emph{low-posdis}. The latter language
is depicted in black in Fig.~\ref{fig:distributions} of the main text. Both languages
come from the training configuration with 2 100-valued input
attributes and 3 100-symbol positions.

\paragraph{Mutual information profiles.} Table \ref{tab:MI} reports mutual information for
the two languages. Note how the highly entangled \emph{low-posdis} is
almost uniform across the table cells.

\begin{table}[tbh]
  \centering
  \begin{tabular}{l|rr|rr}
    \multicolumn{1}{c}{}&\multicolumn{2}{c}{\emph{medium-posdis}}&\multicolumn{2}{c}{\emph{low-posdis}}\\
    &\emph{att1}&\emph{att2}&\emph{att1}&\emph{att2}\\
    \hline
    \emph{pos1}&1.10       &2.01       &1.72       &1.95\\
    \emph{pos2}&0.19       &4.16       &1.74       &1.71\\
    \emph{pos3}&4.44       &0.13       &2.16       &1.77\\
  \end{tabular}
  \caption{Mutual information of each position with each attribute for the studied languages.}
  \label{tab:MI}
\end{table}

\paragraph{Vocabulary usage.} Considering all messages produced after
training for the full training and test set inputs, effective
vocabulary usage for \emph{pos1}, \emph{pos2} and \emph{pos3} are as follows (recall that
100 symbols are maximally available):
\begin{itemize}
\item \emph{medium-posdis}: 91, 96, 98
\item \emph{low-posdis}: 99, 99, 100
\end{itemize}
Although vocabulary usage is high in both cases, \emph{medium-posdis} is
slightly more parsimonious than \emph{low-posdis}.

\paragraph{Ablation studies.} Table \ref{tab:shuffle-results} reports
ablation experiments with both languages. The results for
\emph{medium-posdis} are discussed in the main text. We observe here
how virtually any ablation strongly impacts accuracy in denoting
either attribute by the highly entangled \emph{low-posdis}
language. This points to another possible advantage of (partially)
disentangled languages such as \emph{medium-posdis}: since \emph{pos2}
and \emph{pos3} are referring to \emph{att2} and \emph{att1}
independently, in ablations in which they are untouched, Receiver can
still retrieve partial information, by often successfully guessing the
attribute they each refer to. We also report in the table the effect
of shuffling \emph{across the positions} of each message. This is very
damaging not only for \emph{medium-posdis}, but for \emph{low-posdis}
as well, showing that even the latter is exploiting positional
information, albeit in an inscrutable, highly entangled way. Note in Fig.~\ref{fig:distributions} of the main article that neither language has high \emph{bos}.

\begin{table}[tbh]
  \centering
  \begin{small}
    \begin{tabular}{l@{\,}l@{\,}|@{\,}r@{\,}|@{\,}r@{\,}|@{\,}r@{\,}|@{\,}r@{\,}|@{\,}r@{\,}|@{\,}r@{}}
      \multicolumn{2}{c}{}       &\multicolumn{3}{c}{\emph{medium-posdis}}&\multicolumn{3}{c}{\emph{low-posdis}}\\
                      &           &\emph{att1}&\emph{att2}&\emph{both}&\emph{att1}&\emph{att2}&\emph{both}\\
      \hline
      \emph{fixing}   &\emph{pos1}&1      &3      &0      &4      &5      &0 \\
       \emph{1 position} &\emph{pos2}&1        &68       &0        &4        &4        &0  \\
                      &\emph{pos3}&89       &1        &1        &8        &5        &0  \\
      \hline
      \emph{shuffling}&\emph{pos1}&89       &69       &61       &31       &18       &6  \\     
    \emph{1 position}&\emph{pos2}&100      &3        &3        &30     &25       &8  \\     
                      &\emph{pos3}&1        &100      &1        &15       &20       &3  \\
      \hline
      \emph{shuffling}&\emph{msg}&1         &2        &0        &2        &4        &0  \\
  \end{tabular}
\end{small}
\caption{Feeding shuffled messages from the \emph{medium-posdis} and
  \emph{low-posdis} languages to the corresponding trained
  Receivers. Mean percentage accuracy across 10 random shufflings (standard deviation is always $\approx{} 0$) when: \emph{top}: symbols in all positions but
  one are shuffled across the data-set; \emph{middle}: symbols in a
  single position are shuffled across the data-set; \emph{bottom}: shuffling the symbols within each message (ensuring all symbols move). The data-set
  includes all training and test messages produced by the trained
  Sender and correctly decoded in their original form by Receiver
  ($>$99\% of total messages).}
  \label{tab:shuffle-results}
\end{table}


\subsection{Effect of channel capacity on ease of transmission }
\label{sec:retrain}
Table \ref{tab:retrainingAppendix} replicates the ease-of-transmission
analysis presented in the main text across various channel
capacities. We observe in most cases a significantly positive
correlation, that is even higher (1) for larger Receivers and
(2) for emergent languages with shorter messages (smaller $c_{len}$).

\begin{table*}[t]
\centering
\begin{tabular}{llcccccccccccc}
\toprule
 & & \multicolumn{3}{c}{\emph{posdis}} && \multicolumn{3}{c}{\emph{bosdis}} && \multicolumn{3}{c}{\emph{topsim}}\\
\cmidrule{3-5} \cmidrule{7-9} \cmidrule{11-13}
($c_{len}$, $c_{voc}$) & measure & \makecell{GRU\\(500)} & \makecell{GRU\\(50)} & \makecell{FFN\\(500)} && \makecell{GRU\\(500)} & \makecell{GRU\\(50)} & \makecell{FFN\\(500)} && \makecell{GRU\\(500)} & \makecell{GRU\\(50)} & \makecell{FFN\\(500)} \\
\midrule
\multirow{2}{*}{($3$,$50$)} & \small{Learning Speed} & 0.82 & 0.78 & 0.74 && 0.71 & 0.67 & 0.62 && 0.72 & 0.74& 0.66 \\
  & \small{Generalization} & 0.77 & 0.77 & 0.75 && 0.61 & 0.62 & 0.66 && 0.75 & 0.76& 0.74  \\
\multirow{2}{*}{($4$,$50$)} & \small{Learning Speed} & 0.79 & 0.44 & 0.48 && 0.76  & 0.51 & 0.47 && 0.89 & 0.59& 0.61  \\
  & \small{Generalization} & 0.73 & - & 0.50 && 0.77 & 0.27 & 0.54 && 0.84 & 0.41 & 0.61  \\
\multirow{2}{*}{($6$,$50$)} & \small{Learning Speed} & 0.82 & 0.77 & 0.79 && 0.79 & 0.76 & 0.77 && 0.89& 0.85 & 0.87  \\
  & \small{Generalization} & 0.78 &0.56 & 0.69 && 0.76 &0.55 & 0.67 && 0.85 & 0.65 & 0.77 \\
\multirow{2}{*}{($8$,$50$)} & \small{Learning Speed} & 0.75 & 0.56 & 0.78 && 0.80 & 0.68 & 0.78 && 0.75 & 0.55 & 0.71  \\
  & \small{Generalization} & 0.67 & 0.27 & 0.68 && 0.78 & 0.41 & 0.70 && 0.53 & - & 0.54 \\
\multirow{2}{*}{($10$,$50$)} & \small{Learning Speed} & 0.51 & 0.29 & 0.60 && 0.42 & 0.31 & 0.48 && 0.72 & 0.49& 0.73  \\
  & \small{Generalization} & 0.39& - & 0.44 && 0.47 & - & 0.36 && 0.41 & 0.27& 0.57  \\
\multirow{2}{*}{($12$,$50$)} & \small{Learning Speed} & -& - & - && 0.33 & - & - && 0.49 & - & 0.35  \\
  & \small{Generalization} & - & -0.28 & - && - & - & - && - & - & -\\
\multirow{2}{*}{($3$,$100$)} & \small{Learning Speed} & 0.87 & 0.71 & 0.35 && 0.85 & 0.68 & 0.33 && 0.87 & 0.71& 0.35  \\
  & \small{Generalization} & 0.80 & 0.55 & 0.50 && 0.81& 0.55 & 0.51 && 0.79 & 0.54& 0.48  \\
\multirow{2}{*}{($4$,$100$)} & \small{Learning Speed} & 0.84 & 0.54 & 0.43 && 0.82 & 0.54 & 0.46 && 0.86 & 0.57& 0.49  \\
  & \small{Generalization} & 0.82& 0.38 & 0.47 && 0.80 & 0.39 & 0.47 && 0.82 & 0.41& 0.48  \\
\multirow{2}{*}{($6$,$100$)} & \small{Learning Speed} & 0.88 & 0.83 & 0.80 && 0.89 & 0.78 & 0.78 && 0.94 & 0.87& 0.83  \\
  & \small{Generalization} & 0.87 & 0.68 & 0.68 && 0.90 & 0.69 & 0.67 && 0.90 & 0.70& 0.68  \\
\multirow{2}{*}{($10$,$100$)} & \small{Learning Speed} & 0.85 & 0.58 & 0.62 && 0.82 & 0.59 & 0.64 && 0.72 & 0.74& 0.66  \\
  & \small{Generalization} & 0.86 & 0.39 & 0.47 && 0.81  & 0.50 & 0.37 && 0.72 & 0.35& 0.46  \\
\multirow{2}{*}{($8$,$100$)} & \small{Learning Speed} & 0.73  & 0.58 & 0.65 && 0.79 & 0.59 & 0.65 && 0.70 & 0.57& 0.66  \\
  & \small{Generalization} & 0.69 & 0.39 & 0.37 && 0.67 & 0.37 & 0.37 && 0.49 & 0.34& 0.46  \\
\multirow{2}{*}{($12$,$100$)} & \small{Learning Speed} & 0.39 & - & 0.27 && 0.69 & - & 0.40 && 0.67 & -& 0.51 \\
  & \small{Generalization} & 0.38& - & 0.34 && 0.52 & - & 0.38 && 0.36 & -& -   \\
\hline
\hline
\multirow{2}{*}{Average} & \small{Learning Speed} & 0.75 & 0.62 & 0.61 && 0.72 & 0.63 & 0.59 && 0.79 & 0.67 & 0.62 \\
  & \small{Generalization} & 0.71 & 0.42 & 0.54 && 0.72 & 0.49 & 0.51 && 0.70 & 0.51 & 0.63  \\
\bottomrule
\end{tabular}
\caption{Statistically significant ($p < 0.01$) Spearman correlations between retraining performance (measured by new Receiver Learning Speed and Generalization) and compositionality measures (\emph{posdis}, \emph{bosdis} and \emph{topsim}) for ($i_{att}=2 $, $i_{val}=100$) and different channel capacity. `-' indicates no significant correlations. \label{tab:retrainingAppendix}}
\end{table*}

\end{document}